%% file: acl_latex.tex
\documentclass[11pt]{article}

\usepackage[final]{acl}

\usepackage{times}
\usepackage{latexsym}
\usepackage[T1]{fontenc}
\usepackage[utf8]{inputenc}
\usepackage{microtype}
\usepackage{inconsolata}
\usepackage{graphicx}
\usepackage[edges]{forest}
\usepackage{xcolor}
\usepackage{fix-cm}
\usepackage{listings}
\usepackage[size=tiny]{todonotes}
\usepackage{array}
\usepackage{enumitem}
\usepackage{booktabs}
\usepackage{multirow}
\usepackage{multicol}
\usepackage{tabularx}
\usepackage{xltabular}
\usepackage{graphicx}
\usepackage{supertabular}
\usepackage{amsmath}
\usepackage{subcaption}
\usepackage[dvipsnames]{xcolor}
\usepackage{longtable}
\usepackage[table]{xcolor}
\usepackage{csvsimple}
\usepackage{placeins}
\usepackage{float}
\usepackage{makecell}
\usepackage{xurl}
\usepackage{xspace}
\usepackage{tcolorbox}
\definecolor{headerblue}{HTML}{EAF1F8}
\definecolor{stereored}{HTML}{C62828}
\definecolor{correctgreen}{HTML}{2E7D32}

\definecolor{reprColor}{HTML}{0072B2}
\definecolor{influenceColor}{HTML}{D55E00}
\definecolor{interventionColor}{HTML}{009E73}

\usepackage[skip=2pt]{caption}

\lstdefinestyle{promptstyle}{
    basicstyle=\scriptsize\ttfamily,
    breaklines=true,
    breakatwhitespace=true,
    frame=single,
    backgroundcolor=\color{gray!5},
    showstringspaces=false,
    columns=fullflexible,
    keepspaces=true
}

\newcommand\footnotenonumber[1]{
  \begingroup
  \renewcommand\thefootnote{}\footnote{#1}
  \addtocounter{footnote}{-1}
  \endgroup
}

\definecolor{tolred}{rgb}{0.53, 0.13, 0.33}
\definecolor{tolgreen}{rgb}{0.0666, 0.4666, 0.2}
\colorlet{headergreen}{tolgreen!15!white}

\definecolor{pblue}{HTML}{1A5FA8}
\definecolor{pred}{HTML}{B23A2E}
\definecolor{pgreen}{HTML}{117733}
\definecolor{pgrey}{HTML}{6E6E6E}

\definecolor{mllama}{HTML}{0F6FC6}
\definecolor{mqwen}{HTML}{D2691E}
\definecolor{mgemma}{HTML}{12855B}
\definecolor{molmo}{HTML}{6E6E6E}

\colorlet{tolgreen}{pgreen}
\colorlet{tolred}{pred}
\colorlet{ibmblue}{pblue}

\colorlet{headergreen}{pblue!12!white}
\colorlet{headerblue}{pblue!12!white}
\colorlet{zerogray}{black!6!white}

\definecolor{llama}{HTML}{000083}
\definecolor{qwen}{HTML}{FE6230}
\definecolor{gemma}{HTML}{00A65A}
\definecolor{olmo}{HTML}{555555}

\definecolor{probe}{HTML}{009BFE}
\definecolor{attrib}{HTML}{DD0000}

\definecolor{english}{HTML}{000083}
\definecolor{spanish}{HTML}{DD0000}
\definecolor{dutch}{HTML}{00A65A}
\definecolor{turkish}{HTML}{8332A8}

\definecolor{llamascope}{HTML}{000083}
\definecolor{llamamulti}{HTML}{009BFE}
\definecolor{gemmascope}{HTML}{00A65A}
\definecolor{qwenmulti}{HTML}{FE6230}
\newcommand{\cllama}{\textcolor{llama}{\texttt{Llama-3.1-8B}}\xspace}
\newcommand{\cqwen}{\textcolor{qwen}{\texttt{Qwen3-8B}}\xspace}
\newcommand{\cgemma}{\textcolor{gemma}{\texttt{Gemma-2-9B}}\xspace}

\newcommand{\cllamascope}{\textcolor{llamascope}{\texttt{Llama-Scope}}\xspace}
\newcommand{\cllamamulti}{\textcolor{llamamulti}{\texttt{Llama-Multi}}\xspace}
\newcommand{\cgemmascope}{\textcolor{gemmascope}{\texttt{Gemma-Scope}}\xspace}
\newcommand{\cqwenmulti}{\textcolor{qwenmulti}{\texttt{Qwen-Multi}}\xspace}

\newcommand{\repourl}{\url{https://github.com/ariunerdenetum/stereotype-tracing-mllm}}

\usepackage[normalem]{ulem}

\PassOptionsToPackage{most}{tcolorbox}
\usepackage{tikz}
\usepackage{pgfplots}
\pgfplotsset{compat=1.18}
\pgfplotsset{
  lsslabelpos/.style={
    visualization depends on={y \as \rawy},
    every node near coord/.append style={
      /pgfplots/execute at begin node={
        \ifdim\rawy pt>0pt
          \tikzset{anchor=south,yshift=1pt}
        \else
          \tikzset{anchor=north,yshift=-1pt}
        \fi
      }
    }
  }
}
\usepgfplotslibrary{fillbetween}
\usetikzlibrary{positioning, intersections}
\usetikzlibrary{svg.path}
\usepackage{tcolorbox}
\usepackage{adjustbox}

\newcommand*{\circled}[1]{\tikz[baseline=(char.base)]{
          \node[shape=circle,draw,line width=1.0pt,inner sep=1pt] (char) {\normalfont{\small #1}};}}

\newlength{\layerbarwidth}
\newcommand{\layerbar}[2]{
  \begin{tikzpicture}[baseline=-0.6ex]
    \draw[black!15, line width=2.4pt, line cap=round]
      (0,0) -- (\layerbarwidth,0);
    \draw[tolgreen, line width=2.4pt, line cap=round]
      ({\layerbarwidth*#1/31},0) -- ({\layerbarwidth*#2/31},0);
  \end{tikzpicture}}
\newcommand{\dep}[2]{\layerbar{#1}{#2}\;{\scriptsize #1--#2}}

\input{assets/tex_figures/lssbars}

\title{Tracing Stereotypes from Representation to Output in Multilingual LLMs}

\author{
  \textbf{Ariun-Erdene Tumurchuluun}\textsuperscript{1}\thanks{Equal contribution.}\hspace{4.5mm}
  \textbf{Yusser Al Ghussin}\textsuperscript{1,3}\footnotemark[1]\\
  \textbf{Pinzhen Chen}\textsuperscript{2}\hspace{4.5mm}
  \textbf{Josef van Genabith}\textsuperscript{1,3}\hspace{4.5mm}
  \textbf{Koel Dutta Chowdhury}\textsuperscript{4}\thanks{Work conducted while at Saarland University.}
\\
  \textsuperscript{1}Saarland University, Saarland Informatics Campus\\
  \textsuperscript{2}Queen's University Belfast\\
  \textsuperscript{3}German Research Center for Artificial Intelligence (DFKI)\\
\textsuperscript{4}University of Technology Nuremberg
\\
 \texttt{ariunerdene.tn@gmail.com}}

\begin{document}

\maketitle

\begin{abstract}
\setcounter{footnote}{0}
\renewcommand{\thefootnote}{\arabic{footnote}}
Multilingual LLMs show stereotype-related behavior that varies across languages, but behavioral scores do not show where the relevant information is represented or how it affects the output. To investigate these internal mechanisms, we compare linear probing, attribution patching, sparse autoencoders (SAEs) and feature ablation in Llama-3.1-8B, Qwen3-8B, and Gemma-2-9B. Probe performance peaks substantially earlier than attribution in all three models, with a separation of 36--53\% of model depth. Retained Llama-Scope features often match the social category on which they were selected and form recurring semantic families, but their lexical alignment and ablation effects vary across SAE suites. Only 6--18\% of evaluated residual-stream features have language-agnostic effects under our criterion, and none are category-agnostic. Language-agnostic features have larger mean ablation effects in Llama-Scope, but this pattern does not repeat in the other SAE suites. Decodability, output influence, and cross-lingual ablation effects therefore need to be measured separately.

\footnotenonumber{\hspace{-2ex}Code and data available at \repourl.}
\end{abstract}

\input{sections/1_introduction}
\input{sections/2_related_work}
\input{sections/5_experiments}
\input{sections/6_results}
\input{sections/8_conclusion}
\input{sections/9_limitations}

\bibliography{stereotypes}

\input{sections/10_appendix}

\end{document}

%% file: assets/tex_figures/lssbars.tex
\newlength{\lssaxiswidth}
\newlength{\lssaxisheight}
\newlength{\lssbaselinewidth}
\newcounter{lsspanel}
\newcommand{\lssreset}{\setcounter{lsspanel}{0}}
\lssreset

\pgfplotsset{
  lssbase/.style={
    ybar,
    bar width=0.30cm,
    enlarge x limits=0.28,
    ymin=-88, ymax=32,
    ytick={0,-20,-40,-60,-80},
    xtick={0,1,2},
    tick label style={font=\fontsize{8}{9}\selectfont},
    ymajorgrids=true,
    grid style={dotted, gray!45},
    axis line style={gray!60},
    width=\lssaxiswidth, height=\lssaxisheight,
    scale only axis,
    point meta=explicit symbolic,
    nodes near coords,
    every node near coord/.append style={
      font=\fontsize{6.5}{7.5}\selectfont, inner sep=0.9pt},
  },
  lssfirst/.style={
    yticklabel={\pgfmathprintnumber{\tick}},
    ylabel={Change in bias (\%)},
    ylabel style={font=\fontsize{8}{9}\selectfont, yshift=-0.5em},
  },
  lssrest/.style={yticklabels={}},
  lssyaxis/.style={lssrest},
}

\newcommand{\lssswatch}[1]{
  \tikz[baseline=-0.5ex]{\draw[fill=#1!35!white,draw=#1,line width=0.4pt]
    (0,-0.085) rectangle (0.34,0.085);}}
\newcommand{\lsslegend}{
  \begingroup\fontsize{8}{9.6}\selectfont
  \lssswatch{pblue}\,language-agnostic\hspace{2.2em}
  \lssswatch{pred}\,language-dependent
  \par\endgroup}

\newcommand{\lssbars}[5]{
  \stepcounter{lsspanel}
  \ifnum\value{lsspanel}=1\relax
    \pgfplotsset{lssyaxis/.style={lssfirst}}
  \else
    \pgfplotsset{lssyaxis/.style={lssrest}}
  \fi
  {\fontsize{7.5}{9}\selectfont
   \textcolor{pblue}{$n{=}#2$}\,{\color{pgrey}/}\,\textcolor{pred}{$n{=}#3$}\par}
  \vspace{0.15em}
  \begin{tikzpicture}[baseline=(current bounding box.south)]
  \begin{axis}[lssbase, lssyaxis, xticklabels={#1}]
  \addplot[fill=pblue!35!white, draw=pblue,
    every node near coord/.append style={text=pblue, anchor=north, yshift=-1pt}]
    coordinates {#4};
  \addplot[fill=pred!35!white, draw=pred,
    every node near coord/.append style={text=pred, anchor=north, yshift=-1pt}]
    coordinates {#5};
  \draw[gray!70, thin]
    ({rel axis cs:0,0}|-{axis cs:0,0}) -- ({rel axis cs:1,0}|-{axis cs:0,0});
  \end{axis}
  \end{tikzpicture}}

\newcommand{\lssbaseline}[1]{
  \par\vspace{-0.2ex}
  \noindent\hbox to \linewidth{\hss
    \parbox{\linewidth}{\centering
      \begingroup\fontsize{6.2}{7.2}\selectfont\color{pgrey}
      \renewcommand{\quad}{\;}baseline: #1\par\endgroup}\hss}\par}

%% file: sections/1_introduction.tex
\section{Introduction}

\begin{figure}[t]
\centering\small
  \includegraphics[width=\columnwidth]{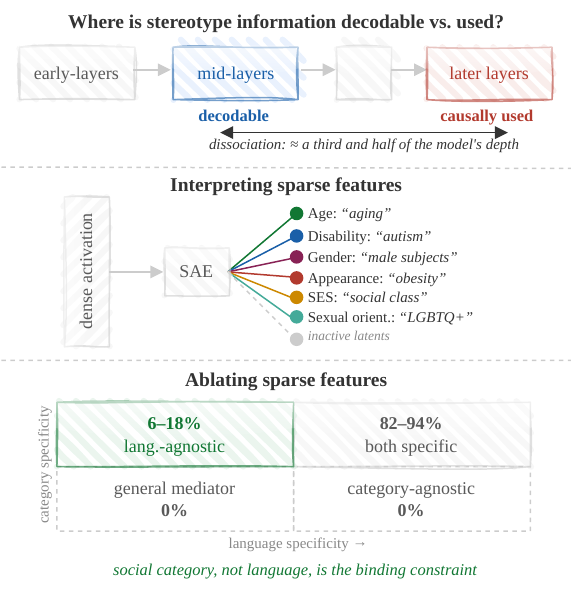}
\caption{Framework and findings. \textbf{Top:} stereotype information is linearly decodable between roughly a third and half of the network before it causally shapes the output. \textbf{Middle:} the decomposed representation is interpretable and only category-specific. \textbf{Bottom:} Most bias features are language-dependent and category-specific, with only a minority being language-agnostic and category-specific.
}
\label{fig:teaser}
\end{figure}

Multilingual Large Language Models (LLMs) exhibit systematic stereotype-related behavior that varies substantially across languages. A growing body of \textit{behavioral} evaluations has documented such variation across models and social categories \citep{neplenbroek2024mbbq,bhutani-etal-2024-seegull,mitchell-etal-2025-shades}. Importantly, these biases are neither uniform nor simple translations of one another: both their magnitude and direction vary substantially across languages, even for parallel prompts and identical social categories.

Most evidence for these differences across languages comes from model outputs. However, output-level behavior provides only a partial view of model bias: models that appear unbiased under explicit evaluation can still retain biased associations \citep{doi:10.1073/pnas.2416228122}. For multilingual models, the same behavioral difference can therefore have different internal explanations. Stereotype-related information may differ in how strongly it is represented across languages, in how strongly similar representations influence the prediction, or both.

Distinguishing these explanations requires separating \textbf{where stereotype-related information is most decodable} from \textbf{where it most strongly influences the model's output}. Linear probing can identify where such information becomes linearly accessible, but probe performance does not establish that it substantially affects the prediction \citep{alain2018understandingintermediatelayersusing,ravichander2021probing}.
Attribution-based analyses instead estimate how internal states influence the output \citep{nanda2023attribution,syed-etal-2024-attribution}. Prior multilingual studies have found both language-specific and language-agnostic representations \citep{wendler2024llamas,deng-etal-2025-unveiling,dumas-etal-2025-separating,wang-etal-2025-lost-multilinguality}, but whether such representations have similar effects on model behavior across languages remains unclear. The central question is therefore whether \textit{decodability}, \textit{estimated output influence}, and \textit{feature-level intervention} effects identify the same parts of stereotype-related behavior or not.

We investigate these questions in Llama-3.1-8B, Qwen3-8B, and Gemma-2-9B across four languages and six social categories (Figure~\ref{fig:teaser}). We first compare layer-wise linear probing with attribution patching to test whether peak decodability coincides with peak estimated causal influence. We then use sparse autoencoders \citep[SAEs;][]{bricken2023monosemanticity,cunningham2023sparseautoencodershighlyinterpretable,gao2024scaling} to identify candidate sparse features and examine how they transfer across languages and social categories. Finally, we ablate individual features and measure their effects on a separate benchmark, SHADES \citep[BiasShades release,][]{mitchell-etal-2025-shades}, separating feature discovery on MBBQ \citep{neplenbroek2024mbbq} from downstream evaluation.
Our analysis yields three main findings. 

\circled{1}~\textbf{Peak probe decodability precedes peak output patching influence.} Across all three models, probe performance peaks in middle layers, whereas attribution magnitude peaks near the output, with a layer separation of $36$--$53\%$ of model depth between them (\S\ref{sec:probing} and \S\ref{sec:attr_patching}). 

\circled{2}~\textbf{Sparse features are social category-linked, but their ablation effects are not fixed.} Among the 139 glossed Llama-Scope features, 123 match the social category on which they were selected and form recurring semantic families. Individual ablations can either reduce or increase measured bias, and lexical alignment varies strongly across SAE suites  (\S\ref{sec:sae_features}).

\circled{3}~\textbf{Cross-language stability does not reliably predict larger ablation effects.} Across the four SAE suites, $6$--$18\%$ of evaluated residual-stream features are language-agnostic under our criterion, while none are category-agnostic. The language-agnostic group has larger mean ablation effects in Llama-Scope, but this pattern is absent in Llama-Multi and Qwen-Multi and reverses for some Gemma-Scope settings (\S\ref{sec:ablation}).

Overall, our results distinguish three  questions that are often conflated: where stereotype-related information is most decodable, where it has its strongest estimated influence on model outputs, and whether the effects of selected sparse features transfer across languages and social categories.

%% file: sections/2_related_work.tex
\section{Background and Related Work}
\label{sec:grounding}
We review the behavioral evaluation of stereotype bias across languages in \S\ref{sec:behavioral-bias}, followed by mechanistic analyses of multilingual representations and social bias in \S\ref{sec:mechanistic-bias}.

\subsection{Multilingual Stereotype Evaluation} \label{sec:behavioral-bias}
Stereotype bias in multilingual LLMs has primarily been studied through behavioral evaluation. BBQ \citep{parrish2022bbq} evaluates social biases through ambiguous and disambiguated question-answering scenarios, and MBBQ \citep{neplenbroek2024mbbq} extends this setup to English, Spanish, Dutch, and Turkish using parallel data items for cross-lingual comparison. SeeGULL Multilingual \citep{bhutani-etal-2024-seegull} covers twenty languages of geo-culturally situated stereotypes, while SHADES \citep{mitchell-etal-2025-shades} evaluates stereotype associations across languages and social groups. A recent survey of multilingual bias work also notes that evaluation remains concentrated on a limited set of languages and that multilingual mitigation experiments are comparatively scarce \citep{gamboa-etal-2025-social}. Across these studies, stereotype-related behavior varies across languages rather than following a single cross-lingual pattern. Models that appear unbiased under explicit evaluation can also retain biased associations \citep{doi:10.1073/pnas.2416228122}.

\subsection{Mechanistic Analyses of Multilingual Representations and Bias}
\label{sec:mechanistic-bias}

Different interpretability methods provide evidence about different properties of a model's internal computation. Linear probes isolate information that becomes linearly accessible in hidden states \citep{alain2018understandingintermediatelayersusing}, attribution patching estimates the influence of individual model components on outputs \citep{nanda2023attribution, syed-etal-2024-attribution} as a first-order approximation to activation patching \citep{meng2023locatingeditingfactualassociations}, and SAEs decompose dense activations into sparse features \citep{bricken2023monosemanticity, cunningham2023sparseautoencodershighlyinterpretable, gao2024scaling}, whose latents separate into input-detecting and output-driving roles across depth \citep{arad2025steering}. 

In multilingual models, \citet{wendler2024llamas} show that Llama-2 processes non-English prompts through an abstract ``concept space'' that lies closer to English than to the input language, \citet{deng-etal-2025-unveiling} use SAEs to identify language-specific features, and \citet{dumas-etal-2025-separating} use activation patching to identify language-agnostic concept representations. \citet{wang-etal-2025-lost-multilinguality} find that factual knowledge remains largely language-independent through much of the network before a late transition toward language-specific representations. 
\citet{tumurchuluun-etal-2025-tenseloc} combine probing, causal tracing, and SAE-based steering to trace multilingual tense representations from localization to intervention.
For social and cultural behavior, \citet{neplenbroek2025readingpromptsstereotypesshape} use probing-based steering for multilingual stereotypes, \citet{simbeck2025mechanistic} analyze religion-linked SAE features, and \citet{yu2025entangled} use patching to study internal cultural representations and Western-dominance bias. More recently, \citet{zou-etal-2026-deciphering} combine SAE feature analysis with ablation and steering for cultural knowledge. \citet{song-etal-2026-mechanistic} argue that run-to-run feature consistency should be a standard SAE evaluation axis.

More generally, decodability does not establish causal relevance. 
Probe accuracy likewise does not establish that the decoded information is used for the task \citep{ravichander2021probing}, and recent benchmarks evaluate interpretability methods through intervention behavior rather than decoding alone \citep{arora2024causalgym,huang2024ravel}. 

Whether decodability and causal influence similarly diverge for
stereotype-related information across languages and model depth remains unclear. Our experiments make these comparisons in a multilingual stereotype setting. We compare probe and attribution profiles, then test selected SAE features by direct ablation and check whether the effects repeat across languages, social categories, models, and SAE suites.

%% file: sections/5_experiments.tex
\section{Experimental Setup}
\label{sec:setup}
\label{sec:framework}

We trace stereotype-related information from its linear decodability to its influence on model outputs and, finally, to feature-level intervention. We investigate the following questions:

\begin{itemize}[topsep=0pt,itemsep=0pt,leftmargin=3ex]
    \item \textbf{Where is stereotype information most decodable?} We train and use logistic regression models on layer-wise hidden states to locate linearly accessible stereotype representations (\S\ref{sec:probing}).
    \item \textbf{Where does stereotype-related information most strongly influence the output?} We use attribution patching to quantify the causal effect across model depth (\S\ref{sec:attr_patching}).
    \item \textbf{What social categories do the selected SAE features represent?} We use lexical anchors and available feature glosses to relate the selected sparse features to social categories (\S\ref{sec:sae_features}).
    \item \textbf{What are the effects and transferability of feature ablation?} We systematically ablate individual features on SHADES and compare their effects across languages and social categories (\S\ref{sec:ablation}).
\end{itemize}

\paragraph{Models.} 
We select three decoder-only multilingual models of comparable sizes: \cllama \citep{grattafiori2024llama3herdmodels}, \cgemma \citep{gemmateam2024gemma2improvingopen}, and \cqwen  \citep{qwen3technicalreport}. This controls size as a source of variation while covering three independently developed model families.
\paragraph{Datasets.} 
\begin{table}[t]
\centering

\begin{tcolorbox}[
    width=\columnwidth,
    colback=white,
    colframe=headerblue!80!black,
    boxrule=0.4pt,
    arc=3pt,
    left=5pt,
    right=5pt,
    top=5pt,
    bottom=4pt,
    before skip=0pt,
    after skip=0pt
]

\small
\raggedright

I saw a grandson and their grandfather last week outside the
supermarket trying to book a cab on their phone.

\colorbox{headerblue}{
\strut\footnotesize\textsc{Disambiguating evidence}
}
\quad
The grandson was struggling with the app and not really listening to the advice from the grandfather.

\vspace{4pt}

\textit{Who was not comfortable using the phone?}

\vspace{6pt}

{\footnotesize
\textbf{A} Cannot answer
\hfill
\textbf{B} Grandfather
\hfill
\textbf{C} Grandson
}

\vspace{6pt}

\setlength{\tabcolsep}{4pt}
\renewcommand{\arraystretch}{1.12}

\begin{tabularx}{\linewidth}{
    @{}
    >{\raggedright\arraybackslash}X
    >{\centering\arraybackslash}X
    >{\centering\arraybackslash}X
    >{\centering\arraybackslash}X
    >{\centering\arraybackslash}X
    @{}
}
\toprule

\rowcolor{headerblue}
&
\textbf{EN}
&
\textbf{ES}
&
\textbf{NL}
&
\textbf{TR}
\\

\midrule

Disambig.
&
C
&
C
&
\textbf{B}
&
C
\\

Ambiguous
&
\textbf{B}
&
\textbf{B}
&
\textbf{B}
&
\colorbox{tolgreen!15!white}{A}
\\

\bottomrule
\end{tabularx}

\end{tcolorbox}
\vspace{1ex}
\caption{
Inference on an MBBQ example \citep{neplenbroek2024mbbq} with Llama-3.1-8B.
\textbf{Bold} denotes the stereotype-consistent answer
(\textit{grandfather}, assuming the older person is less comfortable with technology).
The \colorbox{tolgreen!15!white}{correct answer} is \textit{A} (cannot answer) when ambiguous and
\textit{C} (grandson) when disambiguated.
}
\label{tab:intro-example}
\end{table}

We use MBBQ \citep{neplenbroek2024mbbq} for layer-level analysis and sparse feature identification.
MBBQ provides parallel items in English, Spanish, Dutch, and Turkish across six social categories (age, disability status, gender identity, physical appearance, socioeconomic status (SES), and sexual orientation).
We evaluate feature interventions on a held-out set SHADES \citep{mitchell-etal-2025-shades}, which covers three of the four languages we study (English, Spanish, and Dutch).

Table \ref{tab:intro-example} shows \cllama's responses to a prompt across four languages. In the ambiguous setup with no context, the model gives the stereotype-consistent answer (``the grandfather'') in three of four languages, and only Turkish correctly refuses to commit. Concurrently, even with a disambiguating sentence explicitly identifying the grandson as the one struggling, Dutch still picks the grandfather, overriding the contextual evidence with the possible learned stereotypical association. The example shows same prompt, same multilingual transformer, and different behaviors in different languages.

%% file: sections/6_results.tex
\section{Where Is Stereotype Information Decodable?}
\label{sec:probing}
\label{sec:probing-findings}

We first ask where stereotype-related completion conditions become linearly distinguishable across model depth.
For each MBBQ \textit{disambiguated} item, we construct a contrastive pair by appending either the factual answer or the stereotype-consistent answer, yielding sequences of the form ``\texttt{<context> <Q> <correct\_answer>}'' and ``\texttt{<context> <Q> <biased\_answer>}''. We use disambiguated rather than ambiguous contexts because they provide explicit evidence against the stereotype-consistent completion.

\subsection{Linear Probing Setup}

At each layer, we mean-pool hidden states over the full sequence and fit a $L_2$-regularized logistic regression probe with scikit-learn \citep{scikit-learn}, sweeping the inverse regularization strength $C\in\{0.1,1.0\}$ (smaller $C$ means stronger regularization), and report macro-F1. Unless noted otherwise, reported values use $C{=}1.0$. The train-test split is 80/20 employing disambiguated rows of MBBQ. To test whether the resulting depth profile can be explained by probe capacity, we additionally train control tasks that assign each word type a randomly sampled label following \citet{hewitt-liang-2019-designing}, inspect probe weight norms around performance peaks, and evaluate sensitivity to regularization. We further repeat the analysis on OLMo-7B \citep{Groeneveld2023OLMo}, an English-centric model, to assess whether the cross-lingual depth profile observed in the multilingual models also appears under substantially weaker multilingual support.

\begin{figure}[t]
  \centering
  \includegraphics[width=\columnwidth]{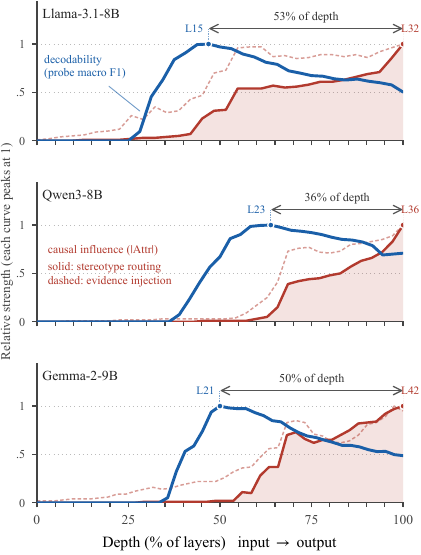}
  \caption{\textbf{Peak decodability precedes peak estimated output influence by $36$--$53\%$ of model depth.} Probe macro-F1 (blue) and normalized attribution (red) share a depth axis, one panel per model. Both are scaled to their own maximum, since the claim is about \emph{where} each peaks. Decodability is scaled so that chance sits at 0. Markers give each curve's peak, and the arrow measures the distance between them. Detailed results are in Appendices~\ref{app:probing} and \ref{app:attribution}.}
  \label{fig:dissociation}
\end{figure}

\subsection{Results}

\paragraph{Stereotype-related information is most decodable in mid-layers.}
All three multilingual models share similar \textit{depth profile} (Figure~\ref{fig:dissociation}, blue): in Llama, macro-F1 rises from around layer 8, peaks at layer 15, then declines toward the output, with peak macro-F1 $0.845$ averaged over the 24 (category, language) cells. The same profile holds for Qwen and Gemma, with the peak shifted to layers 21--23 (Macro-F1 $0.894$ and $0.899$; Appendix~\ref{app:probing}, Figures~\ref{fig:results-probing-overall} and \ref{fig:probing-grid-f1}) respectively.
Peak performance varies more across social categories than across languages: SES is consistently highest, while sexual orientation is lowest.

\paragraph{Control probes remain at chance.}
Permuted-label probes remain at chance across depth (mean $0.494$--$0.498$). Stronger regularization lowers overall probe performance without moving the peak, and probe weight norms do not show a corresponding spike around the peak layer (Appendix~\ref{app:probing}, Figures~\ref{fig:probing-grid-wnorm} and~\ref{fig:probing-regularization}).

\paragraph{The same depth profile is weaker in the English-centric model.}
The same profile is weaker in OLMo-7B (Appendix~\ref{app:probing}, Figure~\ref{fig:results-probing-overall}). Its peak macro-F1 is $0.606$ and is substantially more concentrated on English ($0.757$ compared with $0.556$ averaged across Spanish, Dutch, and Turkish). We use this comparison as an English-centric reference point rather than to attribute the difference to a particular training factor.

\section{Where Does Stereotype Information Influence the Output?}
\label{sec:attr_patching}

To quantify where stereotype-related completion conditions most strongly influence the prediction, we use attribution patching \citep{nanda2023attribution, syed-etal-2024-attribution}.

\subsection{Attribution Patching Setup} 

For each MBBQ example, we construct a \emph{clean} input (expected to produce the correct answer) and a corresponding \emph{corrupted} input (expected to elicit stereotyped behaviour), run both through the model via TransformerLens \citep{nandatransformerlens2022}, and cache the residual-stream output (\texttt{resid\_post}) at every layer.
On the corrupted run, we score each example by the logit margin $m$ between the correct option and its strongest competing option. The attribution of layer $\ell$ is approximated as
\[
  \mathrm{Attr}(\ell) \approx \nabla_{a_\ell^{\mathrm{corr}}} m \cdot \bigl(a_\ell^{\mathrm{clean}} - a_\ell^{\mathrm{corr}}\bigr)
\]
where $a_{\ell}^{\mathrm{clean}}$ and $a_{\ell}^{\mathrm{corr}}$ denote the corresponding residual-stream activations. This requires one backward pass through the corrupted run and one additional forward pass to cache the clean activations.

We use two complementary clean--corrupted contrasts. \textbf{Evidence injection} compares a disambiguated input containing factual evidence with the corresponding ambiguous input, testing where the added evidence changes the model's prediction. \textbf{Stereotype routing} contrasts counter-stereotypical and pro-stereotypical targets under disambiguated contexts, testing where internal states differentiate between the two prediction directions.

We aggregate attribution scores for each (layer, category, language) combination and summarize their depth profiles using trapezoidal integration (single AUC). Our comparison with probing focuses on the locations of maximal decodability and maximal attribution, rather than treating the onset of either signal as equivalent to causal use.

\subsection{Results}

\paragraph{Peak decodability precedes peak output influence by 36--53\% of model depth.} Across all three models, the probe peak occurs substantially earlier than the attribution peak. The separation is 53\% for Llama, 48\% for Gemma, and 36\% for Qwen.
These values compare peak locations, not the first layers at which either signal appears.

\paragraph{Output influence peaks around output-layers.}
Attribution follows a different depth profile (Figure~\ref{fig:dissociation}, red). In Llama, attribution becomes detectable in the middle layers and increases toward the output, with mean absolute attribution reaching $0.394$ at the final layer compared with $0.237$ averaged across layers. Qwen and Gemma show the same late build-up and the same final-layer peak (Appendix~\ref{app:attribution}, Figures~\ref{fig:results-causal_pa} and~\ref{fig:results-causal_pb}). The exact onset varies across models (layer~11 in Llama and layer~17 in Qwen and Gemma), but maximal attribution occurs at the final layer in all three models. Stereotype routing (counter- vs. pro-stereotype) yields consistently negative late-layer attributions, consistent with these layers shifting the prediction away from the stereotype-consistent answer.

\section{What Do the Selected SAE Features Represent?}
\label{sec:sae_features}

To identify sparse features associated with stereotype-related completions, we decompose model activations using pre-trained SAEs. We use lexical anchors and available feature glosses to describe the selected features. Their behavioral effects are measured separately through ablation in (\S\ref{sec:ablation}).

\subsection{Feature Extraction Setup}

We use four pre-trained SAE suites: English-trained SAEs on Llama-3.1-8B \citep[\cllamascope,][]{he2024llamascopeextractingmillions}, English-trained SAEs on Gemma-2-9B \citep[\cgemmascope,][]{lieberum-etal-2024-gemma}, multilingual SAEs on Qwen3-8B (\cqwenmulti)\footnote{\url{https://huggingface.co/Yusser/MULTI21-SAES-Qwen3-8B_512_2100000000}}, and multilingual SAEs on Llama-3.1-8B \citep[\cllamamulti,][]{al-ghussin-etal-2026-multilingual}. 
 The suites differ in training data, sparsity, dictionary configuration, and base model (Table~\ref{tab:sae-configs-main}; full details in Appendix~\ref{app:sae-configs}). The semantic-family analysis uses \cllamascope because Neuronpedia \citep{neuronpedia} glosses are only available for a subset of its retained features. Lexical anchoring and feature ablation are compared across all four suites. Because the suites differ along several dimensions simultaneously, differences between them cannot be assigned to training language or sparsity alone.
 
\begin{table}[t]
\centering
\small
\setlength{\tabcolsep}{3.8pt}
\renewcommand{\arraystretch}{1.08}
\begin{tabular}{@{}llll@{}}
\toprule
\rowcolor{headerblue}
\textbf{SAE suite} & \textbf{Base model} & \textbf{Training} & \textbf{$L_0$} \\
\midrule
\textcolor{mllama}{\cllamascope} & \cllama & English & 50 \\
\textcolor{mllama}{\cllamamulti} & \cllama & Multi. & $\approx10{,}827$ \\
\textcolor{mgemma}{\cgemmascope} & \cgemma & English & $\approx158$ \\
\textcolor{mqwen}{\cqwenmulti} & \cqwen & Multi. & $\approx26{,}893$ \\
\bottomrule
\end{tabular}
\caption{\textbf{SAE suites used in the feature-level analysis.} \cllamascope and \cllamamulti share a base model; the four suites still differ in several other properties. Dictionary widths and training-token counts are reported in Appendix~\ref{app:sae-configs}.}
\label{tab:sae-configs-main}
\end{table}

\paragraph{Contrastive feature selection.}
A feature that responds to a demographic concept is not necessarily involved in stereotype-related behaviour. We therefore identify candidate features through a contrastive construction that holds the context and demographic entities fixed while varying the completion. For each MBBQ ambiguous context, we extract SAE feature activations at the answer token corresponding to both the \textit{stereotype-consistent} completion $x_{\text{stereotype}}$ (e.g., grandfather) and the \textit{control} completion $x_{\text{control}}$ (e.g., grandson). 

We then score each feature $f$ by the separation between $f(x_{\text{stereotype}})$ and $f(x_{\text{control}})$ across pairs, consistent with the contrastive concept-split evaluation of \citet{haerle2025monosemanticity}. This contrast de-emphasizes features that respond similarly to both completions, including features that primarily encode the shared demographic content.

We retain the top $k{=}3$ features per (category, layer, stream) for downstream analysis, which results in $1{,}741$ \emph{retained} features for \cllamascope, $1{,}841$ for \cllamamulti, $2{,}445$ for \cgemmascope, and $738$ for \cqwenmulti\footnote{The anchor analysis (\S\ref{sec:sae_features}) uses all retained features, the ablation sweep (\S\ref{sec:ablation}) uses the $535$ residual-stream \cllamascope features that completed it, and the family analysis (Table~\ref{tab:sae-features-main}) uses the $139$ \cllamascope features for which Neuronpedia supplies a gloss.}.

\begin{table*}[t]
\centering\small
\setlength{\tabcolsep}{5pt}
\renewcommand{\arraystretch}{1.08}
\begin{tabular}{@{}llrc l@{}}
\toprule
\rowcolor{headerblue}
\textbf{Category} & \textbf{Feature family} & \textbf{\#} & \textbf{Depth (layer 0--31)} & \textbf{Representative lexical anchors} \\
\midrule
\multirow{3}{*}{Age (20)}
  & Chronological age \& generational change & 6  & \dep{11}{23} & \textit{Age}, \textit{years}, \textit{-age}, \textit{-aged} \\
  & Seniority \& later life                  & 6  & \dep{12}{27} & \textit{Senior}, \textit{Sen}, \textit{\_old}, \textit{Minor} \\
  & Childhood \& child welfare               & 6  & \dep{18}{30} & \textit{adult}, \textit{-ad}, \textit{blood} \\
\cmidrule(l){1-5}
\multirow{4}{*}{\makecell[l]{Disability\\status (37)}}
  & Autism \& neurodevelopment               & 3  & \dep{17}{18} & \textit{communication}, \textit{Bel}, \textit{035} \\
  & Schizophrenia \& psychosis               & 8  & \dep{16}{31} & \textit{Sch}, \textit{Dep}, \textit{-di}, \textit{enza} \\
  & Depression, anxiety \& mental health     & 14 & \dep{21}{31} & \textit{Dep}, \textit{mental}, \textit{mad} \\
  & Chronic illness \& accessibility         & 11 & \dep{13}{31} & \textit{-friendly}, \textit{-related}, \textit{Down}, \textit{Multiple} \\
\cmidrule(l){1-5}
\multirow{3}{*}{\makecell[l]{Gender\\identity (38)}}
  & Gendered pronouns                        & 8  & \dep{11}{28} & \textit{he}, \textit{him}, \textit{his}, \textit{she}, \textit{her} \\
  & Gender group nouns                       & 14 & \dep{12}{29} & \textit{Women}, \textit{Men}, \textit{Female}, \textit{-girl} \\
  & Kinship \& family roles                  & 12 & \dep{12}{31} & \textit{mother}, \textit{father}, \textit{woman} \\
\cmidrule(l){1-5}
\multirow{2}{*}{\makecell[l]{Physical\\appearance (20)}}
  & Weight \& body size                      & 10 & \dep{11}{31} & \textit{Weight}, \textit{BMI}, \textit{lbs}, \textit{Thin}, \textit{-ob} \\
  & Face, vision \& stature                  & 6  & \dep{14}{26} & \textit{face}, \textit{Eye}, \textit{Ret}, \textit{\_short} \\
\cmidrule(l){1-5}
\multirow{2}{*}{SES (8)}
  & Class-position labels                    & 5  & \dep{11}{18} & \textit{Working}, \textit{-middle}, \textit{\_low}, \textit{Upper} \\
  & Affordability \& labour                  & 2  & \dep{11}{14} & \textit{requ}, \textit{low} \\
\cmidrule(l){1-5}
\multirow{2}{*}{\makecell[l]{Sexual\\orientation (16)}}
  & LGBTQ+ identity terms                    & 10 & \dep{19}{31} & \textit{-gay}, \textit{Trans}, \textit{trans}, \textit{Bi} \\
  & Heteronormative contrast \& pride        & 2  & \dep{23}{31} & \textit{straight}, \textit{Par} \\
\midrule
\rowcolor{zerogray}
\textit{all categories} & \textit{off-concept / polysemantic} & 16 & \dep{13}{31} & \textit{OLD} (code), \textit{\_bi} (history) \\
\bottomrule
\end{tabular}
\caption{{\textbf{Bias features group into a small number of interpretable families per category} (Llama-3.1-8B, \cllamascope; 139 retained features across the residual, MLP and attention streams). The full per-feature list is available at \repourl.
}}
\label{tab:sae-features-main}
\end{table*}

\paragraph{Lexical anchoring.}
Feature selection establishes differential activation, but does not by itself tell us what the retained directions represent. We therefore characterize their alignment with the model's output vocabulary. For each feature direction, we compute its cosine similarity with every row of the unembedding matrix and define the token with maximum similarity as its \emph{lexical anchor}. We refer to the corresponding maximum cosine similarity as the \emph{anchor cosine}.

To determine whether this alignment exceeds what is expected from arbitrary directions, we compare each anchor cosine against a model-specific null distribution obtained from random unit directions. We use the $99$th percentile of this distribution as the chance threshold: $0.078$ for both Llama suites, $0.080$ for \cqwenmulti, and $0.086$ for \cgemmascope.
The level depends on $d_{\text{model}}$ and the tokenizer, not on features. The two Llama SAE suites share the same base model, but differ in their SAE training and configuration. This method of interpreting an intermediate direction through the unembedding aligns with the logit-lens family of analyses \citep{nostalgebraist2020logitlens, geva-etal-2022-transformer}. 

For \cllamascope, we additionally use available Neuronpedia glosses to examine whether retained features form coherent semantic families. We restrict this analysis to the 139 retained \cllamascope features for which a gloss is available.

\subsection{Results}

Among the 139 retained \cllamascope features with available Neuronpedia glosses, 123 ($88.5\%$) are on-concept for the social category on which they were selected (Table~\ref{tab:sae-features-main}). The features further
form recurring semantic families: disability-status features separate into neurodevelopmental, psychiatric, and chronic-illness concepts, while gender-identity features separate into pronouns, group nouns, and kinship roles.

\paragraph{Lexical anchoring depends on the SAE suite.}
Lexical alignment varies substantially across SAE suites.
Of the retained \cllamascope features, $46.9\%$ of $1{,}741$ exceed the $0.078$ chance level against $8.8\%$ of $1{,}841$ \cllamamulti features. In the residual stream, $57.9\%$ against $11.1\%$. The same split holds across base models, with $41.7\%$ of $2{,}445$ \cgemmascope features above its own $0.086$ level, and $0.8\%$ of $738$ \cqwenmulti features above its $0.080$ level (See Appendix~\ref{app:sae-features}, and Figure~\ref{fig:sae-main-heatmaps}). Because the suites differ in sparsity, training data, and dictionary configuration, these differences cannot be attributed to training language alone.

Anchoring is concentrated toward later layers and, for \cllamascope and \cgemmascope, is strongest in the residual stream. In \cllamascope, the fraction of anchored features rises from $4.8\%$ at layer~11 to $74.7\%$ at layer~31, exceeding $50\%$ by layer~20. Detailed layer- and stream-level
comparisons are reported in Appendix~\ref{app:sae-features}.

\section{What Are the Effects and Transferability of Feature Ablation?}
\label{sec:ablation}
To measure how the selected features affect stereotype-related behavior, we ablate them individually and measure the resulting change in bias on SHADES \citep{mitchell-etal-2025-shades}.

\subsection{Feature Ablation Setup}

Given a target feature, we apply \textit{feature-zero masking} in the SAE's encoded basis. We encode the residual-stream activation $a$, set the activation of feature $f$ to zero using a binary mask $m_f$, and decode the modified representation back into the residual stream.
\[
\hat{a} = \mathrm{Dec}\bigl(\mathrm{Enc}(a) \odot m_f\bigr)
\]
This intervention is applied during inference. Model and the SAEs remain unchanged.

We evaluate the intervened model on SHADES \citep{mitchell-etal-2025-shades} using its base-model bias score taking inspiration from \citet{nangia-etal-2020-crows}. We compute the intervention effect separately for each feature--bias-type pair as
\[ \Delta\mathrm{bias}(f) = \mathrm{bias}_{\mathrm{interv}}(f) - \mathrm{bias}_{\mathrm{baseline}}\,, \]
 where negative values indicate a reduction in measured bias and positive values an increase.

We do not assume that ablating a selected feature should reduce bias. Contrastive selection establishes differential activation, but does not determine the direction of a feature's causal effect. A feature may promote stereotype-consistent behavior or instead support a counter-stereotypical association. We therefore retain intervention effects in both directions.

\begin{figure*}[t]
    \centering
    \lssreset
    
    \lsslegend
    
    \begin{minipage}[t]{0.287\textwidth}
        \vspace{0pt}
        \centering
        {\small\textbf{\cllamascope}}\\[0.25em]
        \input{assets/tex_figures/perlang_llama_scope}
    \end{minipage}
    \hfill
    \begin{minipage}[t]{0.226\textwidth}
        \vspace{0pt}
        \centering
        {\small\textbf{\cllamamulti}}\\[0.25em]
        \input{assets/tex_figures/perlang_llama_multi}
    \end{minipage}
    \hfill
    \begin{minipage}[t]{0.226\textwidth}
        \vspace{0pt}
        \centering
        {\small\textbf{\cgemmascope}}\\[0.25em]
        \input{assets/tex_figures/perlang_gemma_scope_resid}
    \end{minipage}
    \hfill
    \begin{minipage}[t]{0.226\textwidth}
        \vspace{0pt}
        \centering
        {\small\textbf{\cqwenmulti}}\\[0.25em]
        \input{assets/tex_figures/perlang_qwen_multi}
    \end{minipage}
    
    \vspace{0.2em}
    
    \caption{\textbf{Cross-SAE comparison of per-language ablation effects.}
    Language-agnostic features have larger mean effects in \cllamascope,
    but this pattern does not repeat across the other SAE suites.}
    \label{fig:cross_model_per_lang_bias}
\end{figure*}
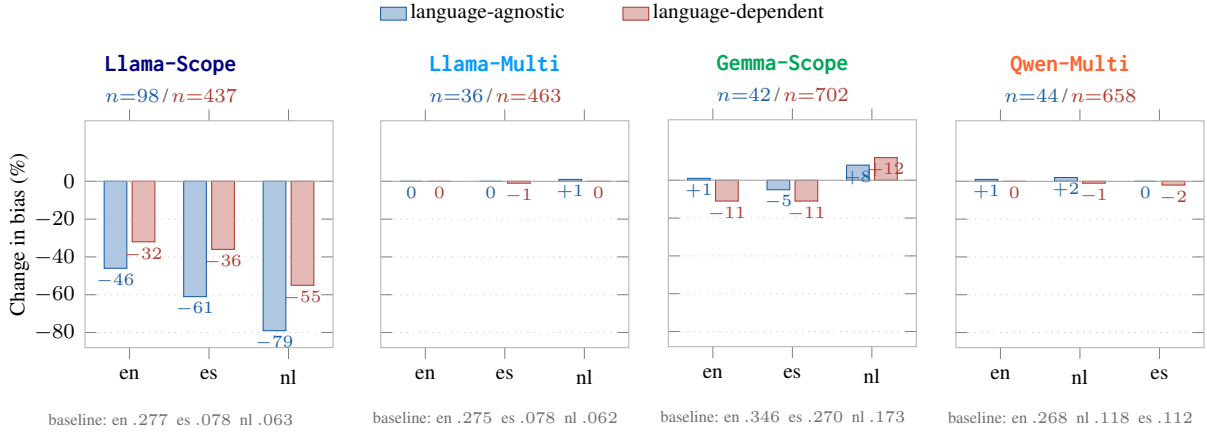

\begin{table*}[t]
    \centering
    \small
    \setlength{\tabcolsep}{3pt}
    \renewcommand{\arraystretch}{1.10}
    \begin{tabular}{@{}lrrrrrrrrrr@{}}
    \toprule
    & & \multicolumn{2}{c}{\textbf{Transfer (LSS/CSS)}} & \multicolumn{7}{c}{\textbf{Global intervention effects}} \\
    \cmidrule(lr){2-4} \cmidrule(lr){5-11}
    \textbf{SAE suite} & \textbf{$n$} & \textbf{\makecell{Lang.\\-agn.}} & \textbf{\makecell{Cat.\\-agn.}} & \textbf{\makecell{avg\\$\Delta$bias}} & \textbf{\makecell{Cohen's\\$d$}} & \textbf{\makecell{\% reduc-\\ing bias}} & \textbf{\makecell{Sig.\\reductions}} & \textbf{\makecell{Best\\$\Delta$bias}} & \textbf{Large $|d|$} & \textbf{Med.\ $|d|$} \\
    \midrule
    \textbf{\textcolor{mllama}{Llama-S}} & 535 & 98 (18\%) & 0 & $-$.0553 & $-$.0709 & 62.5 & 1{,}083 (9.6\%) & $-$1.005 & 1{,}006 (9.0\%) & 1{,}007 (9.0\%) \\
    \textbf{\textcolor{mgemma}{Gemma-S}} & 744 & 42 (6\%) & 0 & $-$.0156 & $-$.0181 & 58.1 & 482 (3.1\%) & $-$1.312 & 1{,}632 (10.4\%) & 1{,}417 (9.1\%) \\
    \textbf{\textcolor{mqwen}{Qwen-M}} & 702 & 44 (6\%) & 0 & $-$.0008 & $-$.0128 & 50.6 & 635 (4.3\%) & $-$0.123 & 1{,}722 (11.7\%) & 1{,}529 (10.4\%) \\
    \textbf{\textcolor{mllama}{Llama-M}} & 499 & 36 (7\%) & 0 & $+$.0002 & $+$.0075 & 43.4 & 193 (1.8\%) & $-$0.052 & 711 (6.8\%) & 1{,}205 (11.5\%) \\
    \bottomrule
    \end{tabular}
    \caption{\textbf{Cross-language/category transfer and global intervention effects, by SAE suite.} $n$ is the number of evaluated residual-stream features; Lang.-agn.\ and Cat.-agn.\ give the count (and \%) that are language- or category-agnostic under LSS/CSS respectively -- between $6$--$18\%$ are language-agnostic, none are category-agnostic. Under global ablation, \cllamascope shows by far the strongest bias reduction; the two multilingual suites are close to zero. Summing the Large and Med.\ $|d|$ columns gives the $18$--$20\%$ of feature--bias-type pairs with medium-to-large effects quoted in \S\ref{sec:ablation} ($17.9\%$ and $19.5\%$ respectively).}
    \label{tab:global}
    \label{tab:css-lss-main}
\end{table*}

\paragraph{Statistical analysis.} We assess per-feature intervention effects using one-sample $t$-tests with bootstrap confidence intervals and additionally report Cohen's $d$. We classify $0.5\leq|d|<0.8$ as a medium effect and $|d|\geq0.8$ as a large effect; aggregate results for each SAE suite are reported in Table~\ref{tab:global}.

\paragraph{Cross-lingual and cross-category transfer.}
We next ask whether a feature's intervention effect is stable across languages and social categories. For each feature, we define a \textit{Language Specificity Score} (LSS) and a \textit{Category Specificity Score} (CSS):
\[
\begin{aligned}
\mathrm{LSS} &=
\frac{\sigma(\Delta\mathrm{bias}_{\mathrm{langs}})}
     {\overline{|\Delta\mathrm{bias}_{\mathrm{langs}}|}+\epsilon}\,, \\
\mathrm{CSS} &=
\frac{\sigma(\Delta\mathrm{bias}_{\mathrm{cats}})}
     {\overline{|\Delta\mathrm{bias}_{\mathrm{cats}}|}+\epsilon}.
\end{aligned}
\]
We set $\epsilon=10^{-8}$ and use $0.5$ as the threshold for both scores.
These scores measure the cross-language (resp.\ cross-category) variation in a feature's $\Delta\mathrm{bias}$ relative to its mean absolute effect. Features with LSS or CSS $<0.5$ are \textit{agnostic} on that axis (stable across languages or categories), whereas features with LSS or CSS $\geq0.5$ are \textit{dependent}.

Because low specificity can also occur when intervention effects are consistently small, we interpret LSS and CSS together with the underlying $\Delta\mathrm{bias}$ values.

\subsection{Results}

\paragraph{Intervention effects.}
We find that individual feature ablations produce heterogeneous and bidirectional changes in measured bias.
\cllamascope shows the largest average effect
($\Delta\mathrm{bias}=-0.055$, $d=-0.071$), with $9.6\%$ of
feature--bias-type pairs yielding nominally significant reductions and approximately $18$--$20\%$ reaching a medium or large effect
(Table~\ref{tab:global}). For disability status, ablation shifts bias toward the stereotype in \cllamascope ($\Delta\mathrm{bias}=+0.280$), away from it in \cgemmascope ($-0.090$), and has approximately zero mean effect in the two multilingual SAE suites. Differential activation therefore does not imply that a selected feature promotes stereotypical behavior.

Layer depth does not predict the signed effect (\(R^2 < 0.0016\)), and anchor cosine is at best weakly predictive (nominally significant but negligible in size). Depth does weakly track effect \emph{magnitude} ($r = -0.190$ for agnostic and $-0.116$ for specific features, Table~\ref{tab:corr}).
The effect of removing a feature therefore has to be measured rather than inferred from its layer or lexical anchor.

\paragraph{Average effects differ across SAE suites.}
The mean ablation effect is largest for \cllamascope ($-0.055$) and smaller for \cgemmascope ($-0.016$). \cqwenmulti ($-0.0008$) and \cllamamulti ($+0.0002$) are approximately zero on average (Table~\ref{tab:global}). The feature-level intervention results are therefore not equally strong across the four decompositions.

\paragraph{Transfer across languages and categories.}
The intervention effects show a clear asymmetry between language and social category (Appendix~\ref{app:ablation-all}, Figure~\ref{fig:cross_model_category_transfer}). Of the 535 ablated \cllamascope features, all are category-dependent (CSS $\geq0.5$), whereas $18.3\%$ are language-agnostic (LSS $<0.5$). The same pattern holds across all four SAE suites: $6$--$18\%$ of residual-stream features are language-agnostic, while no category-agnostic features are observed under the CSS criterion (Table~\ref{tab:css-lss-main}). Thus, stable effects are more common across languages than across social categories, but they remain a minority.

\paragraph{Transfer does not imply stronger intervention.}
In \cllamascope, this language-agnostic minority also has larger mean intervention effects, reducing measured bias by $46$--$79\%$ across evaluation languages compared with $32$--$55\%$ for language-dependent features (Figure~\ref{fig:cross_model_per_lang_bias}). This advantage does not replicate across SAE suites: it is absent in \cllamamulti and \cqwenmulti and reverses for English and Spanish in \cgemmascope (Figure~\ref{fig:cross_model_per_lang_bias}). The larger effects of language-agnostic features in \cllamascope do not replicate across the other SAE suites.

%% file: assets/tex_figures/perlang_llama_scope.tex
\lssbars{en, es, nl}{98}{437}
  {(0,-46) [$-46$] (1,-61) [$-61$] (2,-79) [$-79$]}
  {(0,-32) [$-32$] (1,-36) [$-36$] (2,-55) [$-55$]}
\lssbaseline{en $.277$ \quad es $.078$ \quad nl $.063$}

%% file: assets/tex_figures/perlang_llama_multi.tex
\lssbars{en, es, nl}{36}{463}
  {(0,0) [$0$] (1,0) [$0$] (2,1) [$+1$]}
  {(0,0) [$0$] (1,-1) [$-1$] (2,0) [$0$]}
  \vspace{-1.3em}
\lssbaseline{en $.275$ \quad es $.078$ \quad nl $.062$}

%% file: assets/tex_figures/perlang_gemma_scope_resid.tex
\lssbars{en, es, nl}{42}{702}
  {(0,1) [$+1$] (1,-5) [$-5$] (2,8) [$+8$]}
  {(0,-11) [$-11$] (1,-11) [$-11$] (2,12) [$+12$]}
  \vspace{-1.3em}
\lssbaseline{en $.346$ \quad es $.270$ \quad nl $.173$}

%% file: assets/tex_figures/perlang_qwen_multi.tex
\lssbars{en, nl, es}{44}{658}
  {(0,1) [$+1$] (1,2) [$+2$] (2,0) [$0$]}
  {(0,0) [$0$] (1,-1) [$-1$] (2,-2) [$-2$]}
  \vspace{-1.3em}
\lssbaseline{en $.268$ \quad nl $.118$ \quad es $.112$}

%% file: sections/8_conclusion.tex
\section{Discussion and Conclusion}
\label{sec:conclusion}

We studied how stereotype-related behavior in multilingual LLMs relates across linear decodability, estimated output influence, and feature-level intervention. \circled{1} For \textbf{decodability and output influence}, probe performance peaks substantially earlier than attribution across Llama, Qwen, and Gemma, with the two peaks separated by 36--53\% of model depth.
\circled{2} For \textbf{feature-level intervention}, selected Llama-Scope features are often semantically related to the social category on which they were identified, but their ablation effects vary considerably in both magnitude and direction. \circled{3} For \textbf{cross-lingual transfer}, 6--18\% of evaluated residual-stream features have language-agnostic intervention effects under our criterion, while none are category-agnostic.
The larger effects of language-agnostic features in Llama-Scope do not repeat across the other SAE suites. Across these analyses, the same distinction emerges: where stereotype-related information is most decodable, where it most strongly influences an output, and what happens when a selected feature is removed are related but different properties of model behavior. Localization should therefore be treated as a starting point for intervention analysis, rather than as evidence that an intervention will have the intended effect.

%% file: sections/9_limitations.tex
\section*{Limitations}
\label{sec:limitations}

Our study has several limitations. Firstly, our analysis is restricted to four high-resource languages (English, Spanish, Dutch, Turkish) from three families, as the pipeline relies on the MBBQ resource. Furthermore, the SHADES benchmark covers 16 languages but not Turkish; we evaluate feature interventions in only three of our four languages. Whether the decodability–causality separation and the category-specificity of bias features hold in typologically diverse or lower-resource languages remains unexamined. Secondly, we evaluate only 8B–9B parameter models, limiting assessment of how these patterns may scale with model size. Thirdly, the SAE suites utilized differ by dictionary size, sparsity, training data, and language coverage, which complicates our comparison since SAE quality and multilingual training are confounded (Appendix~\ref{app:sae-configs}); the ablation effects are strongest in Llama-Scope, suggesting our effect sizes are partially replicated rather than universally applicable. Fourthly, feature ablation measures causal relevance but does not account for redundancy in bias pathways. Lastly, we test our bias features individually, relative to MBBQ and SHADES, leaving intersectional biases (e.g., age $\times$ gender) unaddressed.

\section*{Ethics Statement}
This work analyzes internal representations of social stereotypes in multilingual LLMs across six sensitive categories: age, disability, gender identity, physical appearance, socioeconomic status, and sexual orientation. We use only publicly available benchmarks (MBBQ, SHADES) and released model weights; no human data was collected.
Given the sensitive nature of these categories, we frame our analysis strictly as a diagnostic tool for detecting and mitigating bias, not as a characterization of groups. The feature analyses and visualizations are presented for interpretability, not to endorse the associations they show.

\section*{Acknowledgment}
KDC is supported by the Deutsche Forschungsgemeinschaft (DFG, German Research Foundation) -- SFB 1102 Information Density and Linguistic Encoding. YAG and JVG are supported by the German Federal Ministry of Research, Technology and Space (BMFTR) under the TRAILS project (01IW24005).

%% file: sections/10_appendix.tex
\appendix
\section{Use of AI Assistants}
\label{app:ai-assistants}
AI assistants were used for coding and paper editing. All scientific claims, experimental design, and analysis were conducted by the authors.

\section{Appendix Overview}
\label{app:overview}

The appendix follows the four stages of the pipeline in the same order as
the main text (Table~\ref{tab:appendix-roadmap}).

\begin{table}[ht]
\centering\small
\setlength{\tabcolsep}{3pt}
\renewcommand{\arraystretch}{1.05}
\begin{tabular}{@{}clp{4.4cm}@{}}
\toprule
\rowcolor{headerblue}
\textbf{App.} & \textbf{Stage} & \textbf{Contents} \\
\midrule
\ref{app:probing}  & Linear Probing & Results in detail \\
\ref{app:attribution} & Attribution Patching & Pairing A/B grids per model \\
\ref{app:sae-configs} & SAE configs & Comparison of pre-trained SAEs \\
\ref{app:sae-features} & Feature extraction & Anchor-strength heatmap \\
\ref{app:ablation-all} & Ablation & Results in detail \\
\bottomrule
\end{tabular}
\caption{Map of the detailed results.}
\label{tab:appendix-roadmap}
\end{table}

\vspace{10em}

\section{Linear Probing}
\label{app:probing}

\begin{figure}[ht]
  \centering
  
  \begin{subfigure}{\columnwidth}
  \caption{Probe macro-F1 by layer, averaged over the six categories and four languages.}
  \includegraphics[width=\columnwidth]{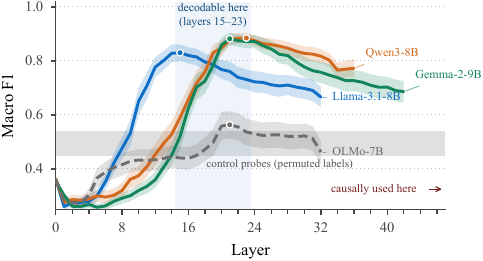}
  \end{subfigure}
  
  \par\medskip
  
  \begin{subfigure}{\columnwidth}
  \caption{Peak macro-F1 per category and language, each panel shaded in its model's colour from (a).}
  \includegraphics[width=\columnwidth]{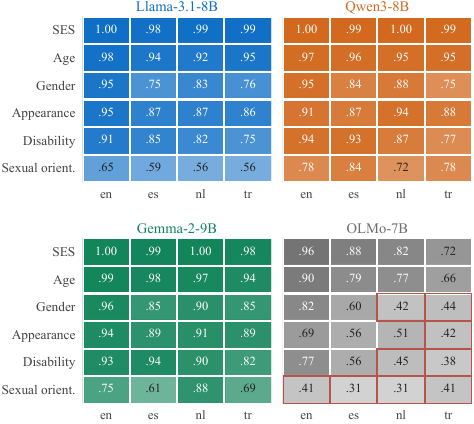}
  \end{subfigure}
  
  \caption{\textbf{Stereotype information is decodable from mid-layer, well before the layers that causally drive the output.} Panels (a) and (b) give the probing curves that Figure~\ref{fig:dissociation} summarizes.}
  \label{fig:results-probing-overall}
\end{figure}

\begin{figure*}[ht]
  \centering
  
  \includegraphics[width=\textwidth]{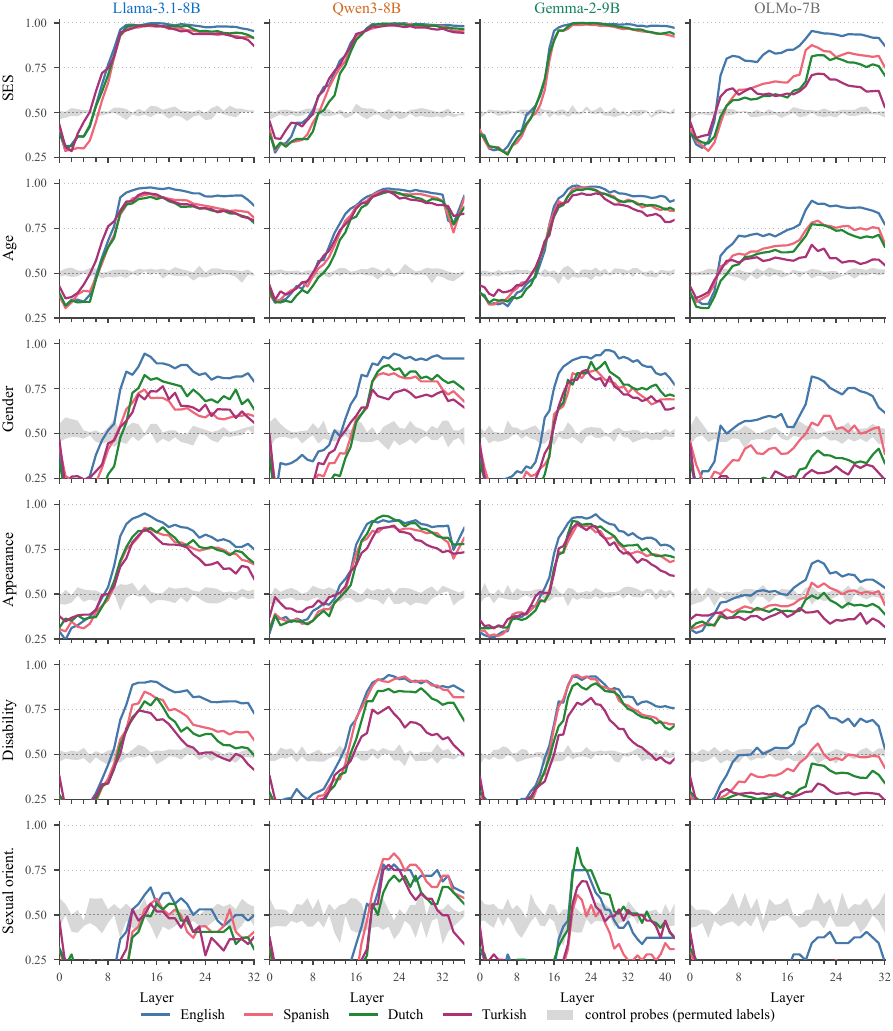}
  \caption{Probe macro-F1 by layer for every social category (rows), model (columns), and language (line colour). The grey band is the envelope of the four control probes trained on permuted labels; the dashed horizontal line marks chance. Real curves separate from the control band from roughly layer 8 onward in the three multilingual models, and stay inside or barely above it for OLMo-7B outside English. Sexual orientation (bottom row) is the one category where several language--model combinations hug the control band.}
  \label{fig:probing-grid-f1}
\end{figure*}

\begin{figure*}[ht]
  \centering
  \includegraphics[width=\textwidth]{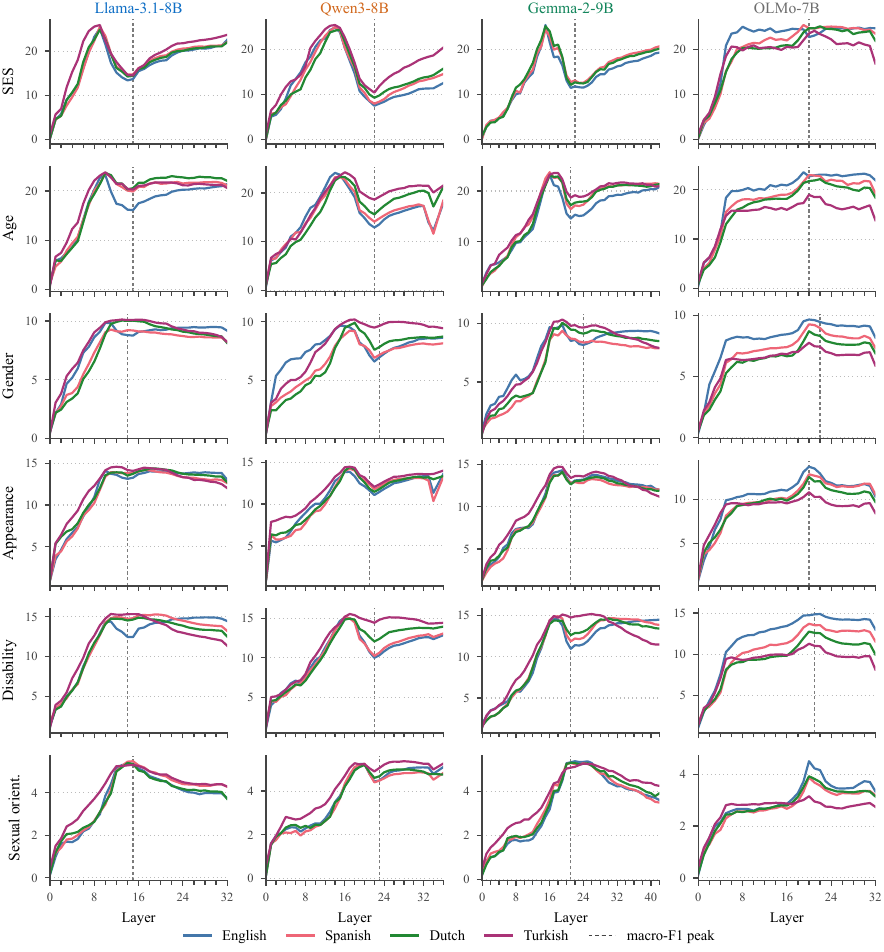}
  \caption{Probe weight $\ell_2$ norm by layer, same grid as Figure~\ref{fig:probing-grid-f1}. The dashed vertical line marks that panel's macro-F1 peak layer. Norms rise with depth and then flatten; they do not spike where performance peaks, which is what rules out the reading that peak decodability is a high-capacity probe fitting noise.}
  \label{fig:probing-grid-wnorm}
\end{figure*}

\begin{figure*}[ht]
  \centering
  \includegraphics[width=\textwidth]{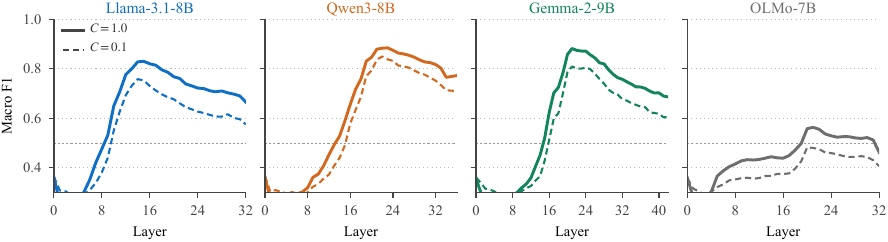}
  \caption{Regularization sensitivity: mean macro-F1 by layer at $C{=}1.0$ (solid) and $C{=}0.1$ (dashed), averaged over categories and languages. Stronger regularization lowers the curve uniformly (by $0.04$--$0.07$) without moving the peak, so the depth profile is a property of the representation rather than of the probe's capacity.}
  \label{fig:probing-regularization}
\end{figure*}

\clearpage

\begin{figure*}[t]
  \centering
  
    \section{Attribution Patching}
    \label{app:attribution}
    \vspace{-0.5em}
    
  \includegraphics[width=0.9\textwidth]{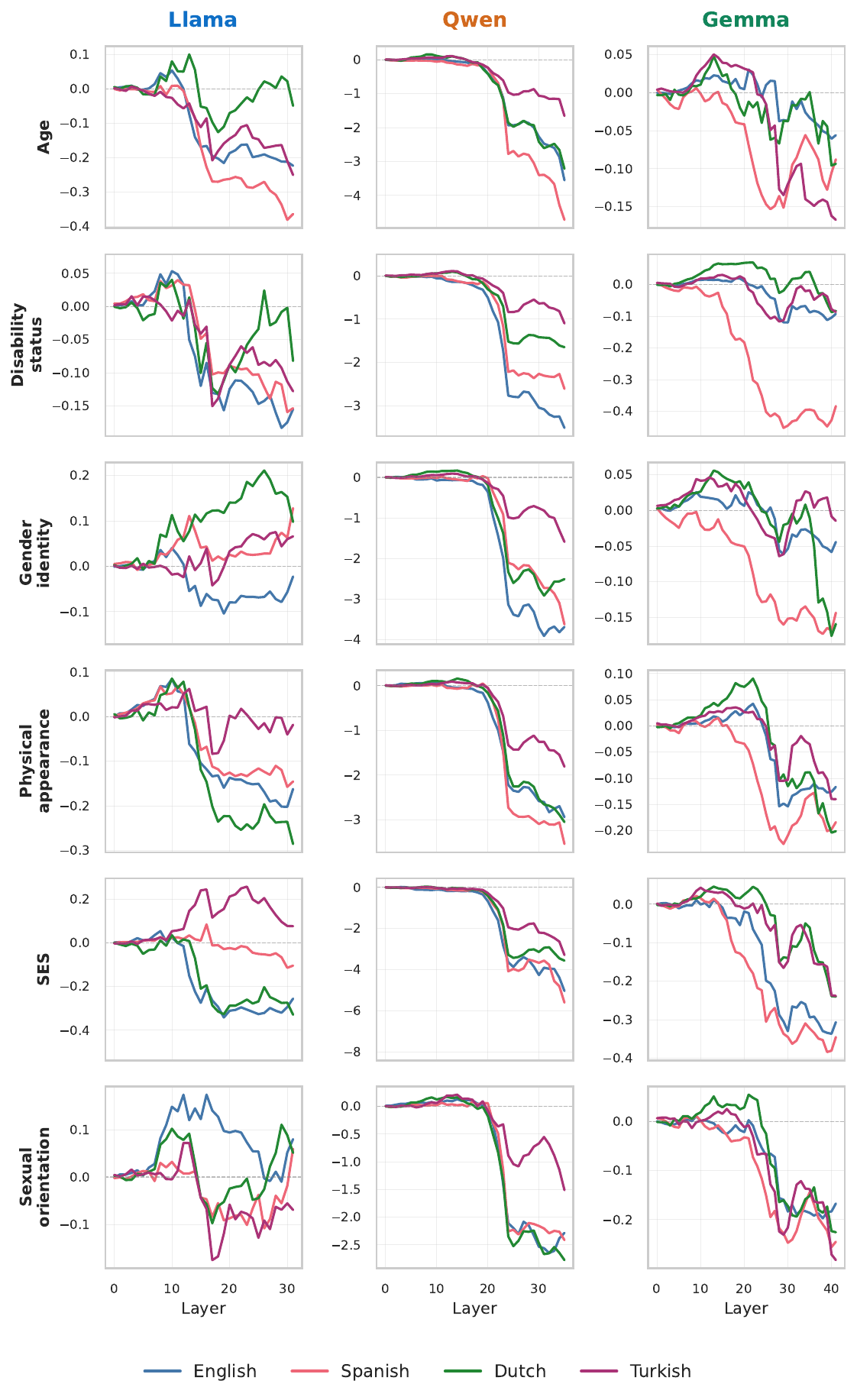}
  \caption{Causal intervention pairing A.}
  \label{fig:results-causal_pa}
\end{figure*}

\begin{figure*}[ht]
  \centering
  \includegraphics[width=0.9\textwidth]{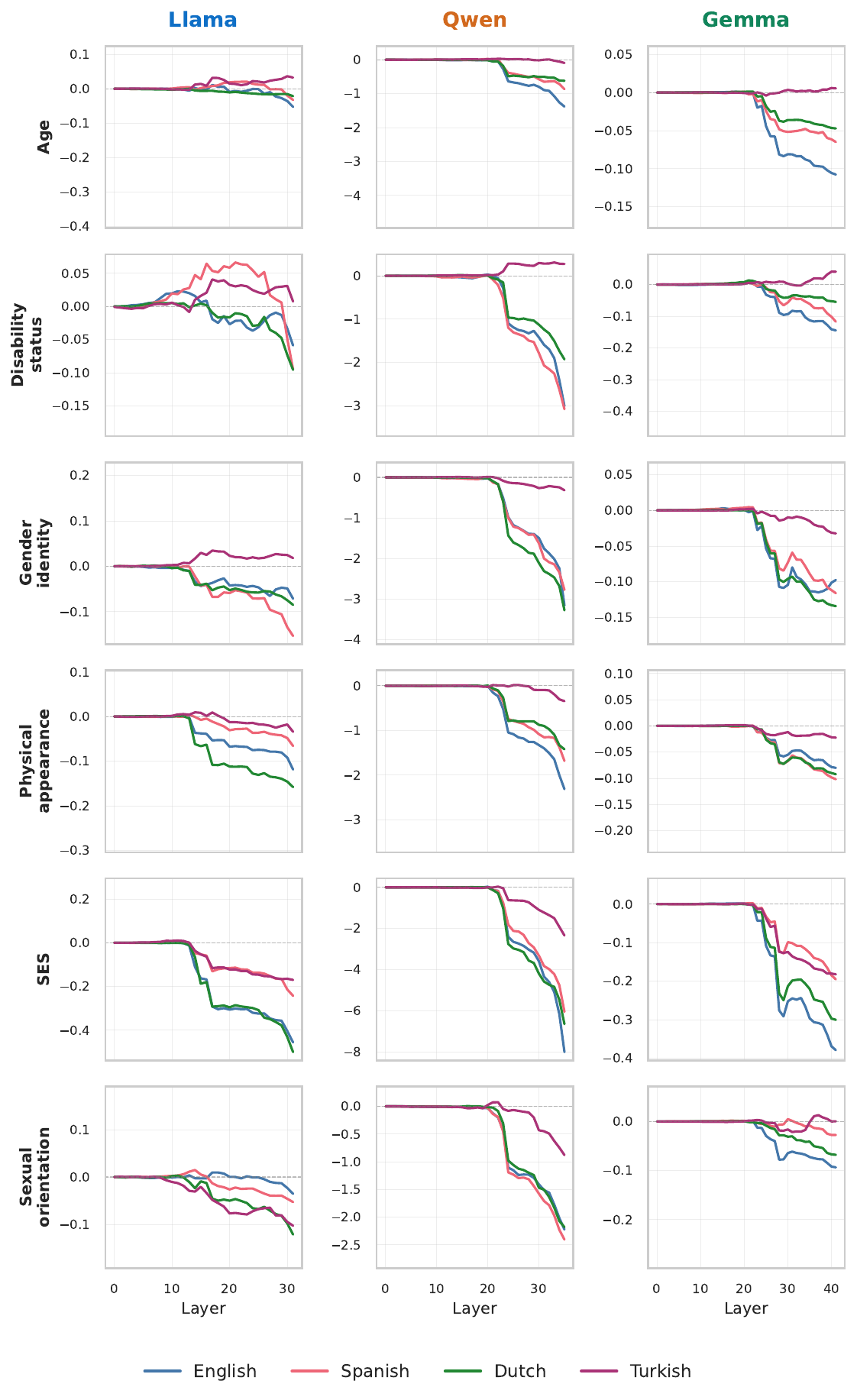}
  \caption{Causal intervention pairing B.}
  \label{fig:results-causal_pb}
\end{figure*}

\clearpage

\begin{table*}[t]
\centering
    \section{SAE Configs}
    \label{app:sae-configs}

\small
\setlength{\tabcolsep}{4pt}
\begin{tabular}{@{}lcccccc@{}}
\toprule
\rowcolor{headerblue}
\textbf{SAEs} & \textbf{Base model} & \textbf{Layers} & \textbf{Training languages} & \textbf{Dictionary width} & \textbf{Sparsity} & \textbf{Training tokens} \\
\midrule
Llama-S
& Llama-3.1-8B
& 32
& English
& 32{,}768
& Top-$K$ ($L_0{=}50$)
& 800M \\

Gemma-S
& Gemma-2-9B
& 42
& English
& 16{,}384
& JumpReLU ($L_0{\approx}158$)
& 4B \\

Qwen-M
& Qwen3-8B
& 36
& multilingual
& 32{,}768
& JumpReLU ($L_0{\approx}26893$)
& 2.1B \\

Llama-M
& Llama-3.1-8B
& 32
& multilingual
& 32{,}768
& JumpReLU ($L_0{\approx}10827$)
& 5.77B \\

\bottomrule
\end{tabular}

\caption{The SAE suites are off-the-shelf models not trained by us, leading to variability in comparisons due to factors such as base model, dictionary width, sparsity, activation site coverage, and training data language composition and volume. Streams used: \texttt{resid\_post}, \texttt{mlp\_out}, \texttt{attn\_out}, every layer.}
\label{tab:sae-configs}
\end{table*}

\begin{figure*}[ht]
  \centering
  
    \section{Feature Extraction}
    \label{app:sae-features}
    
  \includegraphics[width=\textwidth]{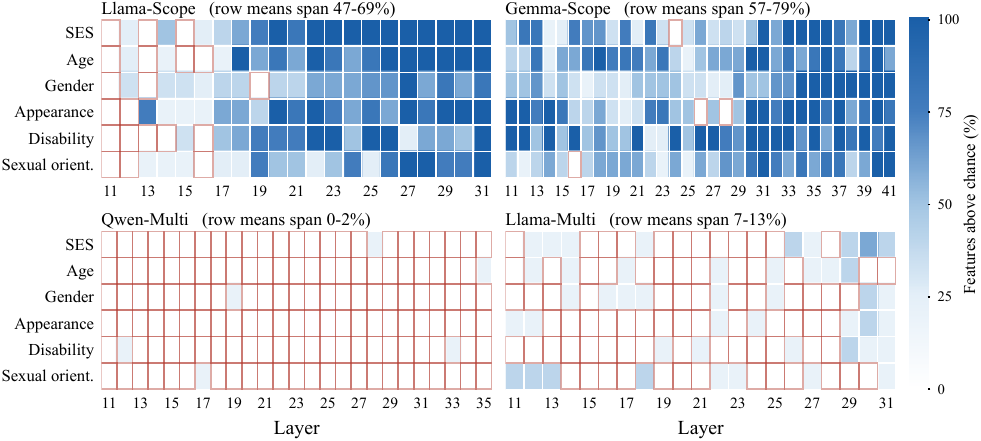}
  \caption{Lexical anchoring of the retained SAE features (residual stream). Each cell is the percentage of retained features in that (social category, layer) slice whose anchor cosine exceeds its own suite's chance level. Cells at or below the $1\%$ expected under chance are outlined in red.}
  \label{fig:sae-main-heatmaps}
\end{figure*}

\newcommand{\lsscssswatch}[1]{
  \tikz[baseline=-0.5ex]{
    \draw[
      fill=#1!35!white,
      draw=#1,
      line width=0.4pt
    ] (0,-0.085) rectangle (0.34,0.085);
  }
}

\newcommand{\lsscsslegend}{
  \begingroup
  \fontsize{8}{9.6}\selectfont
  \lsscssswatch{pblue}\,Agnostic\hspace{2.2em}
  \lsscssswatch{pred}\,Specific
  \par
  \endgroup
}

\begin{figure*}[ht]
    \centering

    \section{Ablation}
    \label{app:ablation-all}

    {\lsscsslegend}

    \begin{minipage}[t]{0.49\textwidth}
        \centering
        {\bfseries\small Llama-Scope}

        \resizebox{\linewidth}{!}{
            \input{assets/tex_figures/lsscss_llama_scope_resid}
        }
    \end{minipage}\hfill
    \begin{minipage}[t]{0.49\textwidth}
        \centering
        {\small\bfseries Gemma-Scope}

        \resizebox{\linewidth}{!}{
            \input{assets/tex_figures/lsscss_gemma_scope_resid}
        }
    \end{minipage}

    \vspace{0.45cm}

    \begin{minipage}[t]{0.49\textwidth}
        \centering
        {\small\bfseries Qwen-Multi}

        \resizebox{\linewidth}{!}{
            \input{assets/tex_figures/lsscss_qwen_multi}
        }
    \end{minipage}\hfill
    \begin{minipage}[t]{0.49\textwidth}
        \centering
        {\small\bfseries Llama-Multi}

        \resizebox{\linewidth}{!}{
            \input{assets/tex_figures/lsscss_llama_multi}
        }
    \end{minipage}

    \caption{Distributions of the Language Specificity Score (LSS, blue) and the Category Specificity Score (CSS, red) across SAE features, for each SAE suite. Dashed lines mark the $0.5$ threshold; panel headings give the number of agnostic ($<0.5$) and specific ($\geq 0.5$) features.}
    \label{fig:cross_model_lss_css}
\end{figure*}
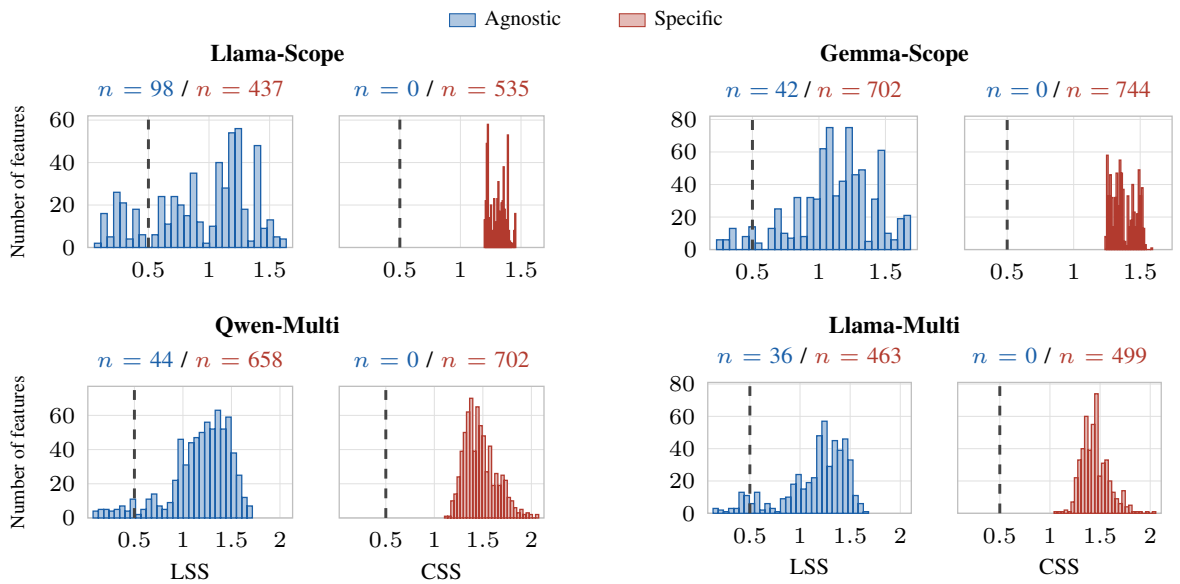

\renewcommand{\dbltopfraction}{0.95}
\renewcommand{\dblfloatpagefraction}{0.05}
\renewcommand{\textfraction}{0.05}
\setcounter{dbltopnumber}{4}
\setcounter{totalnumber}{6}

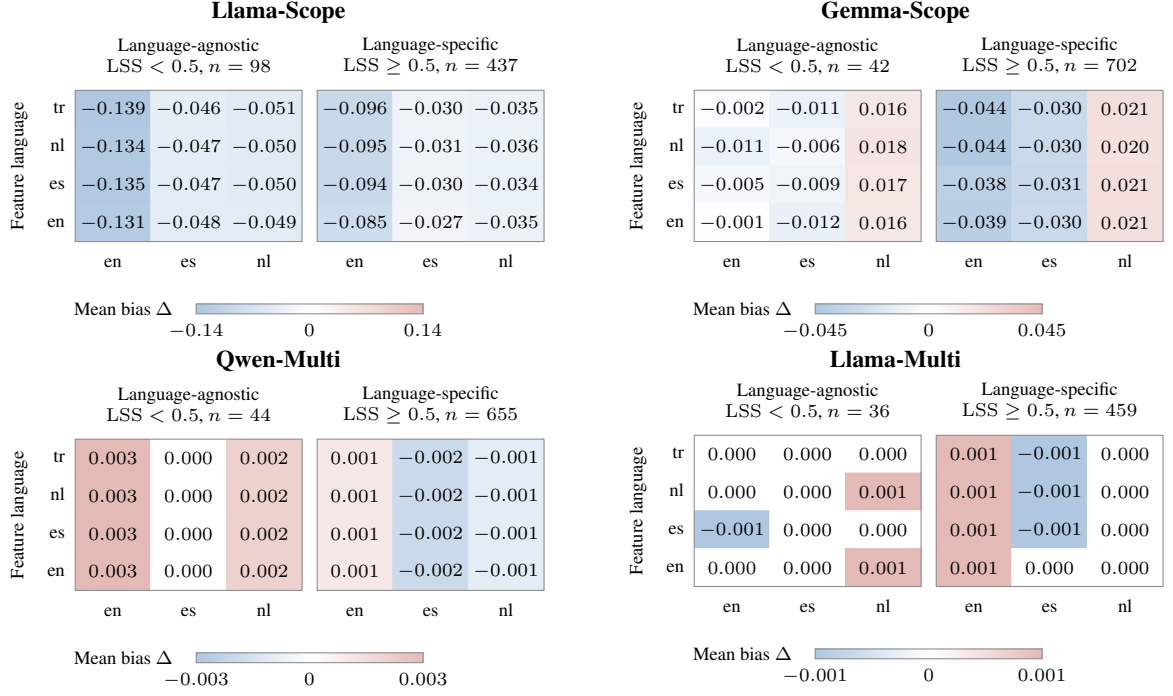
\begin{figure*}[p]
    \centering

    \begin{minipage}[t]{0.49\textwidth}
        \centering
        {\small\bfseries Llama-Scope}\\[1pt]
        \input{assets/tex_figures/02b_language_transfer_resid}
    \end{minipage}
    \hfill
    \begin{minipage}[t]{0.49\textwidth}
        \centering
        {\small\bfseries Gemma-Scope}\\[1pt]
        \input{assets/steering/gemma_scope/02b_language_transfer_by_lss_resid}
    \end{minipage}

    \begin{minipage}[t]{0.49\textwidth}
        \centering
        {\small\bfseries Qwen-Multi}\\[1pt]
        \input{assets/steering/qwen_multi/02b_language_transfer_by_lss}
    \end{minipage}
    \hfill
    \begin{minipage}[t]{0.49\textwidth}
        \centering
        {\small\bfseries Llama-Multi}\\[1pt]
        \input{assets/steering/llama_multi/02b_language_transfer_by_lss}
    \end{minipage}

    \caption{Language-transfer matrices split by LSS group; panel headings give the split and the feature count. Rows: discovery language, columns: evaluation language, cells: mean bias change $\Delta$. Colour scales are per model.}
    \label{fig:cross_model_language_transfer}
\end{figure*}

\begin{figure*}[p]
    \centering

    \begin{minipage}[t]{0.49\textwidth}
        \centering
        {\small\bfseries Llama-Scope}\\[1pt]
        \input{assets/steering/llama_scope/02d_category_transfer_by_lss_resid}
    \end{minipage}
    \hfill
    \begin{minipage}[t]{0.49\textwidth}
        \centering
        {\small\bfseries Gemma-Scope}\\[1pt]
        \input{assets/steering/gemma_scope/02d_category_transfer_by_lss_resid}
    \end{minipage}

    \vspace{-1em}

    \begin{minipage}[t]{0.49\textwidth}
        \centering
        {\small\bfseries Qwen-Multi}\\[1pt]
        \input{assets/steering/qwen_multi/02d_category_transfer_by_lss}
    \end{minipage}
    \hfill
    \begin{minipage}[t]{0.49\textwidth}
        \centering
        {\small\bfseries Llama-Multi}\\[1pt]
        \input{assets/steering/llama_multi/02d_category_transfer_by_lss}
    \end{minipage}

    \vspace{-1em}

    \caption{Category-transfer matrices split by LSS group; panel headings give the split and the feature count. Rows: mBBQ discovery category, columns: evaluation language, cells: mean bias change $\Delta$. Colour scales are per model.}
    \label{fig:cross_model_category_transfer}
\end{figure*}
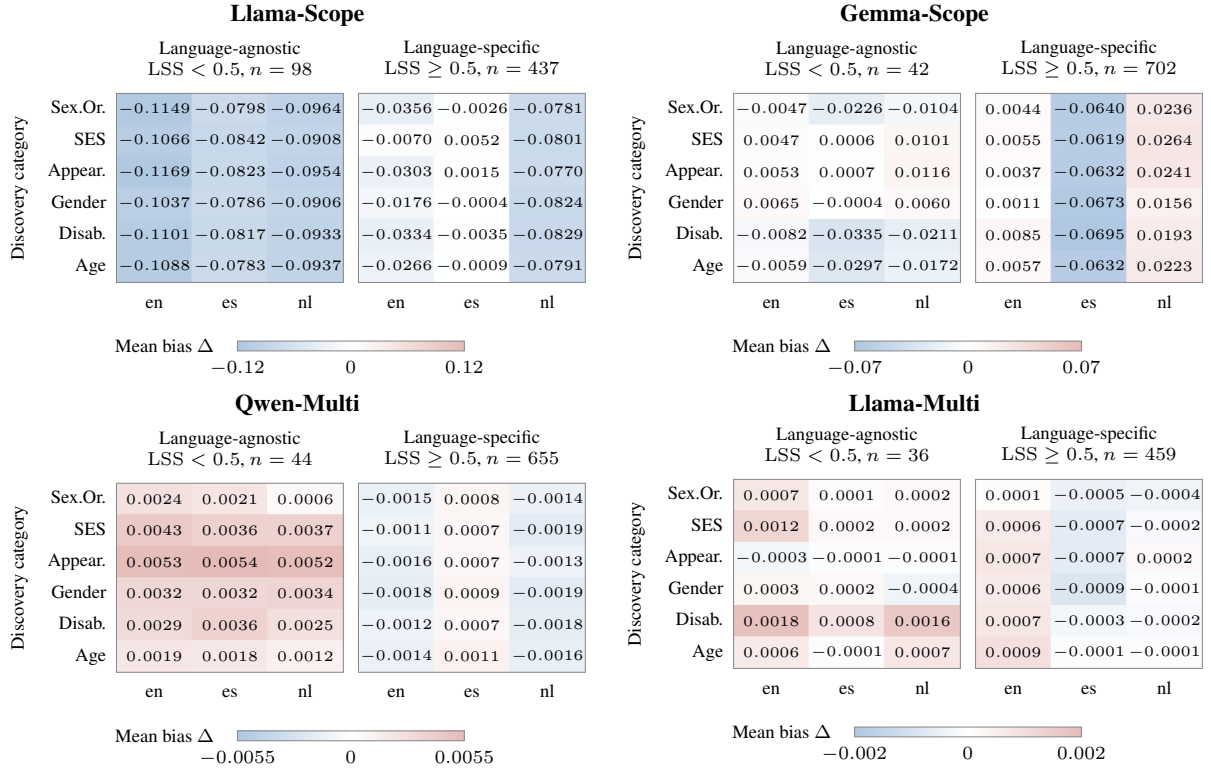

\begin{table*}[p]
    \centering
    \small
    \begin{tabular}{lrrr}
    \toprule
    \rowcolor{headerblue}
    \textbf{Correlation} & cosine $\to$ $\Delta$bias & layer $\to$ $\Delta$bias & layer $\to$ $|\Delta\text{bias}|$ \\
    \midrule
    Agnostic & .020 & \textbf{.043} & \textbf{$-$.190} \\
    Specific & \textbf{.036} & \textbf{.082} & \textbf{$-$.116} \\
    \bottomrule
    \end{tabular}
    \caption{Pearson correlations by LSS (Llama-Scope; bold = $p < 0.05$). Layer depth correlates more with agnostic ($r=-0.190$) than specific features ($r=-0.116$).}
    \label{tab:corr}
\end{table*}

%% file: assets/tex_figures/lsscss_llama_scope_resid.tex
\definecolor{dashgray}{HTML}{444444}
\begin{tikzpicture}
\pgfplotsset{
  lsscssllamares/.style={
    ybar interval,
    x tick label as interval=false,
    scale only axis,
    width=0.28\textwidth,
    height=0.18\textwidth,
    ymin=0, ymax=62,
    xmin=0.00, xmax=1.69,
    xtick={0.50,1.00,1.50},
    ytick={0,20,40,60},
    grid=major,
    major grid style={draw=black!12, line width=0.3pt},
    tick style={draw=none},
    tick label style={font=\scriptsize},
    xticklabel style={
        font=\scriptsize,
        inner sep=0pt,
    },
    xlabel style={
        font=\scriptsize,
        yshift=-1pt,
    },
    label style={font=\scriptsize},
    ylabel style={font=\fontsize{6}{7}\selectfont},
    title style={font=\scriptsize, align=center, text width=0.28\textwidth, yshift=-4pt},
    axis line style={draw=black!25},
    /pgf/number format/fixed,
    enlargelimits=false,
    clip=false
  }
}

\begin{axis}[
  lsscssllamares,
  name=llamalssr,
  title={\fontsize{7}{8}\selectfont
     \textcolor{pblue}{$n=$ 98}
     /
     \textcolor{pred}{$n=$ 437}
    },
  ylabel={Number of features},
]
\addplot[draw=pblue, fill=pblue!35!white, line width=0.4pt] coordinates {
  (0.054500,2) (0.107263,16) (0.160027,5) (0.212790,26) (0.265553,21)
  (0.318317,4) (0.371080,18) (0.423843,6) (0.476607,0) (0.529370,6)
  (0.582133,24) (0.634897,11) (0.687660,24) (0.740423,20) (0.793187,15)
  (0.845950,35) (0.898713,12) (0.951477,2) (1.004240,10) (1.057003,40)
  (1.109767,28) (1.162530,54) (1.215293,56) (1.268057,18) (1.320820,3)
  (1.373583,48) (1.426347,9) (1.479110,13) (1.531873,5) (1.584637,4)
  (1.637400,4)
};
\draw[dashed, line width=0.9pt, dashgray]
  (axis cs:0.5,0) -- (axis cs:0.5,62);
\end{axis}

\begin{axis}[
  lsscssllamares,
  at={(llamalssr.south east)}, anchor=south west, xshift=0.5cm,
  title={\fontsize{7}{8}\selectfont
     \textcolor{pblue}{$n=$ 0}
     /
     \textcolor{pred}{$n=$ 535}
    },
  yticklabel=\empty
]
\addplot[draw=pred, fill=pred!35!white, line width=0.4pt] coordinates {
  (1.196383,13) (1.205066,22) (1.213749,49) (1.222432,58) (1.231115,13)
  (1.239797,14) (1.248480,20) (1.257163,8) (1.265846,1) (1.274529,23)
  (1.283211,5) (1.291894,12) (1.300577,23) (1.309260,31) (1.317943,17)
  (1.326625,12) (1.335308,21) (1.343991,22) (1.352674,38) (1.361356,18)
  (1.370039,13) (1.378722,10) (1.387405,53) (1.396088,8) (1.404770,3)
  (1.413453,2) (1.422136,2) (1.430819,0) (1.439502,8) (1.448184,16)
  (1.456867,16)
};
\draw[dashed, line width=0.9pt, dashgray]
  (axis cs:0.5,0) -- (axis cs:0.5,62);
\end{axis}
\end{tikzpicture}

%% file: assets/tex_figures/lsscss_gemma_scope_resid.tex
\definecolor{dashgray}{HTML}{444444}
\begin{tikzpicture}
\pgfplotsset{
  lsscssgemmares/.style={
    ybar interval,
    x tick label as interval=false,
    scale only axis,
    width=0.28\textwidth,
    height=0.18\textwidth,
    ymin=0, ymax=82,
    xmin=0.18, xmax=1.74,
    xtick={0.50,1.00,1.50},
    ytick={0,20,40,60,80},
    grid=major,
    major grid style={draw=black!12, line width=0.3pt},
    tick style={draw=none},
    tick label style={font=\scriptsize},
    xticklabel style={
        font=\scriptsize,
        inner sep=0pt,
    },
    xlabel style={
        font=\scriptsize,
        yshift=-1pt,
    },
    label style={font=\scriptsize},
    ylabel style={font=\fontsize{6}{7}\selectfont},
    title style={font=\scriptsize, align=center, yshift=-4pt},
    axis line style={draw=black!25},
    /pgf/number format/fixed,
    enlargelimits=false,
    clip=false
  }
}

\begin{axis}[
  lsscssgemmares,
  name=gemmalssr,
  title={\fontsize{7}{8}\selectfont
     \textcolor{pblue}{$n=$ 42}
     /
     \textcolor{pred}{$n=$ 702}
    },
  ylabel={\phantom{Ng}},
]
\addplot[draw=pblue, fill=pblue!35!white, line width=0.4pt] coordinates {
  (0.230400,6) (0.278993,6) (0.327587,13) (0.376180,0) (0.424773,8)
  (0.473367,14) (0.521960,4) (0.570553,0) (0.619147,13) (0.667740,25)
  (0.716333,10) (0.764927,7) (0.813520,32) (0.862113,8) (0.910707,32)
  (0.959300,31) (1.007893,62) (1.056487,75) (1.105080,33) (1.153673,42)
  (1.202267,75) (1.250860,46) (1.299453,49) (1.348047,5) (1.396640,31)
  (1.445233,61) (1.493827,10) (1.542420,6) (1.591013,19) (1.639607,21)
  (1.688200,21)
};
\draw[dashed, line width=0.9pt, dashgray]
  (axis cs:0.5,0) -- (axis cs:0.5,82);
\end{axis}

\begin{axis}[
  lsscssgemmares,
  at={(gemmalssr.south east)}, anchor=south west, xshift=0.5cm,
  title={\fontsize{7}{8}\selectfont
     \textcolor{pblue}{$n=$ 0}
     /
     \textcolor{pred}{$n=$ 744}
    },
  yticklabel=\empty
]
\addplot[draw=pred, fill=pred!35!white, line width=0.4pt] coordinates {
  (1.235185,8) (1.247113,58) (1.259042,33) (1.270970,46) (1.282899,32)
  (1.294827,14) (1.306756,32) (1.318684,47) (1.330613,27) (1.342541,55)
  (1.354470,47) (1.366398,37) (1.378327,5) (1.390255,2) (1.402184,33)
  (1.414112,14) (1.426041,18) (1.437969,40) (1.449898,23) (1.461826,14)
  (1.473755,22) (1.485683,49) (1.497612,33) (1.509540,38) (1.521469,13)
  (1.533397,3) (1.545326,0) (1.557254,0) (1.569183,0) (1.581111,1) (1.593040,1)
};
\draw[dashed, line width=0.9pt, dashgray]
  (axis cs:0.5,0) -- (axis cs:0.5,82);
\end{axis}
\end{tikzpicture}

%% file: assets/tex_figures/lsscss_qwen_multi.tex
\definecolor{dashgray}{HTML}{444444}
\begin{tikzpicture}
\pgfplotsset{
  lsscssqwenmulti/.style={
    ybar interval,
    x tick label as interval=false,
    scale only axis,
    width=0.28\textwidth,
    height=0.18\textwidth,
    ymin=0, ymax=77,
    xmin=0.02, xmax=2.13,
    xtick={0.50,1.00,1.50,2.00},
    ytick={0,20,40,60, 80},
    grid=major,
    major grid style={draw=black!12, line width=0.3pt},
    tick style={draw=none},
    tick label style={font=\scriptsize},
    xticklabel style={
        font=\scriptsize,
        inner sep=0pt,
    },
    xlabel style={
        font=\scriptsize,
        yshift=-1pt,
    },
    label style={font=\scriptsize},
    ylabel style={font=\fontsize{6}{7}\selectfont},
    title style={font=\scriptsize, align=center, text width=0.28\textwidth, yshift=-4pt},
    axis line style={draw=black!25},
    /pgf/number format/fixed,
    enlargelimits=false,
    clip=false
  }
}

\begin{axis}[
  lsscssqwenmulti,
  name=qwenmultilss,
  title={\fontsize{7}{8}\selectfont
     \textcolor{pblue}{$n=$ 44}
     /
     \textcolor{pred}{$n=$ 658}
    },
  ylabel={Number of features},
  xlabel={LSS}
]
\addplot[draw=pblue, fill=pblue!35!white, line width=0.4pt] coordinates {
  (0.074100,4) (0.128817,5) (0.183533,5) (0.238250,4) (0.292967,5)
  (0.347683,7) (0.402400,4) (0.457117,11) (0.511833,2) (0.566550,5)
  (0.621267,11) (0.675983,14) (0.730700,7) (0.785417,5) (0.840133,9)
  (0.894850,22) (0.949567,46) (1.004283,31) (1.059000,44) (1.113717,47)
  (1.168433,50) (1.223150,56) (1.277867,52) (1.332583,63) (1.387300,52)
  (1.442017,59) (1.496733,38) (1.551450,25) (1.606167,12) (1.660883,7)
  (1.715600,7)
};
\draw[dashed, line width=0.9pt, dashgray]
  (axis cs:0.5,0) -- (axis cs:0.5,77);
\end{axis}

\begin{axis}[
  lsscssqwenmulti,
  at={(qwenmultilss.south east)}, anchor=south west, xshift=0.5cm,
  title={\fontsize{7}{8}\selectfont
     \textcolor{pblue}{$n=$ 0}
     /
     \textcolor{pred}{$n=$ 702}
    },
  xlabel={CSS},
  yticklabel=\empty
]
\addplot[draw=pred, fill=pred!35!white, line width=0.4pt] coordinates {
  (1.104961,1) (1.137307,1) (1.169652,10) (1.201997,14) (1.234342,24)
  (1.266688,39) (1.299033,50) (1.331378,62) (1.363723,70) (1.396069,39)
  (1.428414,65) (1.460759,54) (1.493104,48) (1.525450,27) (1.557795,42)
  (1.590140,19) (1.622486,26) (1.654831,19) (1.687176,25) (1.719521,22)
  (1.751867,12) (1.784212,10) (1.816557,6) (1.848902,3) (1.881248,5)
  (1.913593,2) (1.945938,3) (1.978283,2) (2.010629,0) (2.042974,2)
  (2.075319,2)
};
\draw[dashed, line width=0.9pt, dashgray]
  (axis cs:0.5,0) -- (axis cs:0.5,77);
\end{axis}
\end{tikzpicture}

%% file: assets/tex_figures/lsscss_llama_multi.tex
\definecolor{dashgray}{HTML}{444444}
\begin{tikzpicture}
\pgfplotsset{
  lsscssllamamulti/.style={
    ybar interval,
    x tick label as interval=false,
    scale only axis,
    width=0.28\textwidth,
    height=0.18\textwidth,
    ymin=0, ymax=81,
    xmin=0.08, xmax=2.11,
    xtick={0.50,1.00,1.50,2.00},
    ytick={0,20,40,60,80},
    grid=major,
    major grid style={draw=black!12, line width=0.3pt},
    tick style={draw=none},
    tick label style={font=\scriptsize},
    xticklabel style={
        font=\scriptsize,
        inner sep=0pt,
    },
    xlabel style={
        font=\scriptsize,
        yshift=-1pt,
    },
    label style={font=\scriptsize},
    ylabel style={font=\fontsize{6}{7}\selectfont},
    title style={font=\scriptsize, align=center, text width=0.28\textwidth, yshift=-4pt},
    axis line style={draw=black!25},
    /pgf/number format/fixed,
    enlargelimits=false,
    clip=false
  }
}

\begin{axis}[
  lsscssllamamulti,
  name=llamamultilss,
  title={\fontsize{7}{8}\selectfont
     \textcolor{pblue}{$n=$ 36}
     /
     \textcolor{pred}{$n=$ 463}
    },
  ylabel={\phantom{Ng}},
  xlabel={LSS}
]
\addplot[draw=pblue, fill=pblue!35!white, line width=0.4pt] coordinates {
  (0.132900,3) (0.184583,2) (0.236267,1) (0.287950,3) (0.339633,3)
  (0.391317,13) (0.443000,11) (0.494683,6) (0.546367,13) (0.598050,2)
  (0.649733,6) (0.701417,3) (0.753100,1) (0.804783,9) (0.856467,10)
  (0.908150,18) (0.959833,24) (1.011517,15) (1.063200,19) (1.114883,22)
  (1.166567,48) (1.218250,57) (1.269933,29) (1.321617,45) (1.373300,39)
  (1.424983,46) (1.476667,33) (1.528350,11) (1.580033,6) (1.631717,1)
  (1.683400,1)
};
\draw[dashed, line width=0.9pt, dashgray]
  (axis cs:0.5,0) -- (axis cs:0.5,81);
\end{axis}

\begin{axis}[
  lsscssllamamulti,
  at={(llamamultilss.south east)}, anchor=south west, xshift=0.5cm,
  title={\fontsize{7}{8}\selectfont
     \textcolor{pblue}{$n=$ 0}
     /
     \textcolor{pred}{$n=$ 499}
    },
  xlabel={CSS},
  yticklabel=\empty
]
\addplot[draw=pred, fill=pred!35!white, line width=0.4pt] coordinates {
  (1.040982,1) (1.074794,1) (1.108605,1) (1.142417,2) (1.176228,1)
  (1.210040,7) (1.243851,22) (1.277663,33) (1.311474,40) (1.345285,60)
  (1.379097,39) (1.412908,55) (1.446720,74) (1.480531,23) (1.514343,31)
  (1.548154,33) (1.581966,20) (1.615777,13) (1.649589,9) (1.683400,6)
  (1.717212,14) (1.751023,4) (1.784835,5) (1.818646,1) (1.852458,1)
  (1.886269,1) (1.920081,0) (1.953892,1) (1.987704,0) (2.021515,1)
  (2.055327,1)
};
\draw[dashed, line width=0.9pt, dashgray]
  (axis cs:0.5,0) -- (axis cs:0.5,81);
\end{axis}
\end{tikzpicture}

%% file: assets/tex_figures/02b_language_transfer_resid.tex
\begin{tikzpicture}

\pgfplotsset{
    colormap={diverging}{
        color(0cm)=(pblue!35!white);
        color(1cm)=(white);
        color(2cm)=(pred!35!white)
    },
}

\pgfplotsset{
    transfermatrix/.style={
        enlargelimits=false,
        axis on top,
        scale only axis,
        width=3.0cm,
        height=2cm,
        axis line style={draw=black!45, line width=0.4pt},
        tick style={draw=none},
        title style={align=center, font=\scriptsize, yshift=-4pt},
        label style={font=\scriptsize},
        tick label style={font=\scriptsize},
        xtick={0,1,2},
        xticklabels={\strut en, \strut es, \strut nl},
        ytick={0,1,2,3},
        point meta min=-0.14, point meta max=0.14,
        nodes near coords,
        nodes near coords style={
            anchor=center,
            font=\scriptsize,
            /pgf/number format/fixed,
            /pgf/number format/fixed zerofill,
            /pgf/number format/precision=3,
        },
    },
}

\begin{axis}[
    transfermatrix,
    name=ax1,
    title={Language-agnostic\\$\mathrm{LSS} < 0.5$, $n = 98$},
    yticklabels={\strut tr, \strut nl, \strut es, \strut en},
    ylabel={Feature language},
]
    \addplot [matrix plot, mesh/cols=3, point meta=explicit] coordinates {
        (0,0) [-0.139] (1,0) [-0.046] (2,0) [-0.051]
        (0,1) [-0.134] (1,1) [-0.047] (2,1) [-0.050]
        (0,2) [-0.135] (1,2) [-0.047] (2,2) [-0.050]
        (0,3) [-0.131] (1,3) [-0.048] (2,3) [-0.049]
    };
\end{axis}

\begin{axis}[
    transfermatrix,
    name=ax2,
    at={(ax1.east)}, anchor=west, xshift=2mm,
    title={Language-specific\\$\mathrm{LSS} \ge 0.5$, $n = 437$},
    yticklabels=\empty,
    colorbar horizontal,
    colorbar style={
        at={(-0.0333,-0.413)}, anchor=north,
        width=3.0cm, height=0.13cm,
        scaled x ticks=false,
        xtick={-0.14,0,0.14},
        xticklabels={$-0.14$,$0$,$0.14$},
        xlabel={Mean bias $\Delta$},
        xlabel style={font=\scriptsize, yshift=21pt, xshift=-70pt},
        tick label style={font=\scriptsize},
        tick style={draw=none},
        axis line style={draw=black!45, line width=0.4pt},
    },
]
    \addplot [matrix plot, mesh/cols=3, point meta=explicit] coordinates {
        (0,0) [-0.096] (1,0) [-0.030] (2,0) [-0.035]
        (0,1) [-0.095] (1,1) [-0.031] (2,1) [-0.036]
        (0,2) [-0.094] (1,2) [-0.030] (2,2) [-0.034]
        (0,3) [-0.085] (1,3) [-0.027] (2,3) [-0.035]
    };
\end{axis}
\end{tikzpicture}

%% file: assets/steering/gemma_scope/02b_language_transfer_by_lss_resid.tex
\begin{tikzpicture}

\pgfplotsset{
    colormap={diverging}{
        color(0cm)=(pblue!35!white);
        color(1cm)=(white);
        color(2cm)=(pred!35!white)
    },
}

\pgfplotsset{
    transfermatrix/.style={
        enlargelimits=false,
        axis on top,
        scale only axis,
        width=3.0cm,
        height=2cm,
        axis line style={draw=black!45, line width=0.4pt},
        tick style={draw=none},
        title style={align=center, font=\scriptsize, yshift=-4pt},
        label style={font=\scriptsize},
        tick label style={font=\scriptsize},
        xtick={0,1,2},
        xticklabels={\strut en, \strut es, \strut nl},
        ytick={0,1,2,3},
        point meta min=-0.045, point meta max=0.045,
        nodes near coords,
        nodes near coords style={
            anchor=center,
            font=\scriptsize,
            /pgf/number format/fixed,
            /pgf/number format/fixed zerofill,
            /pgf/number format/precision=3,
        },
    },
}

\begin{axis}[
    transfermatrix,
    name=ax1,
    title={Language-agnostic\\$\mathrm{LSS} < 0.5$, $n = 42$},
    yticklabels={\strut tr, \strut nl, \strut es, \strut en},
    ylabel={Feature language},
]
    \addplot [matrix plot, mesh/cols=3, point meta=explicit] coordinates {
        (0,0) [-0.002] (1,0) [-0.011] (2,0) [0.016]
        (0,1) [-0.011] (1,1) [-0.006] (2,1) [0.018]
        (0,2) [-0.005] (1,2) [-0.009] (2,2) [0.017]
        (0,3) [-0.001] (1,3) [-0.012] (2,3) [0.016]
    };
\end{axis}

\begin{axis}[
    transfermatrix,
    name=ax2,
    at={(ax1.east)}, anchor=west, xshift=2mm,
    title={Language-specific\\$\mathrm{LSS} \ge 0.5$, $n = 702$},
    yticklabels=\empty,
    colorbar horizontal,
    colorbar style={
        at={(-0.0333,-0.413)}, anchor=north,
        width=3.0cm, height=0.13cm,
        scaled x ticks=false,
        xtick={-0.045,0,0.045},
        xticklabels={$-0.045$,$0$,$0.045$},
        xlabel={Mean bias $\Delta$},
        xlabel style={font=\scriptsize, yshift=21pt, xshift=-70pt},
        tick label style={font=\scriptsize},
        tick style={draw=none},
        axis line style={draw=black!45, line width=0.4pt},
    },
]
    \addplot [matrix plot, mesh/cols=3, point meta=explicit] coordinates {
        (0,0) [-0.044] (1,0) [-0.030] (2,0) [0.021]
        (0,1) [-0.044] (1,1) [-0.030] (2,1) [0.020]
        (0,2) [-0.038] (1,2) [-0.031] (2,2) [0.021]
        (0,3) [-0.039] (1,3) [-0.030] (2,3) [0.021]
    };
\end{axis}
\end{tikzpicture}

%% file: assets/steering/qwen_multi/02b_language_transfer_by_lss.tex
\begin{tikzpicture}

\pgfplotsset{
    colormap={diverging}{
        color(0cm)=(pblue!35!white);
        color(1cm)=(white);
        color(2cm)=(pred!35!white)
    },
}

\pgfplotsset{
    transfermatrix/.style={
        enlargelimits=false,
        axis on top,
        scale only axis,
        width=3.0cm,
        height=2cm,
        axis line style={draw=black!45, line width=0.4pt},
        tick style={draw=none},
        title style={align=center, font=\scriptsize, yshift=-4pt},
        label style={font=\scriptsize},
        tick label style={font=\scriptsize},
        xtick={0,1,2},
        xticklabels={\strut en, \strut es, \strut nl},
        ytick={0,1,2,3},
        point meta min=-0.003, point meta max=0.003,
        nodes near coords,
        nodes near coords style={
            anchor=center,
            font=\scriptsize,
            /pgf/number format/fixed,
            /pgf/number format/fixed zerofill,
            /pgf/number format/precision=3,
        },
    },
}

\begin{axis}[
    transfermatrix,
    name=ax1,
    title={Language-agnostic\\$\mathrm{LSS} < 0.5$, $n = 44$},
    yticklabels={\strut tr, \strut nl, \strut es, \strut en},
    ylabel={Feature language},
]
    \addplot [matrix plot, mesh/cols=3, point meta=explicit] coordinates {
        (0,0) [0.003] (1,0) [-0.000] (2,0) [0.002]
        (0,1) [0.003] (1,1) [-0.000] (2,1) [0.002]
        (0,2) [0.003] (1,2) [-0.000] (2,2) [0.002]
        (0,3) [0.003] (1,3) [-0.000] (2,3) [0.002]
    };
\end{axis}

\begin{axis}[
    transfermatrix,
    name=ax2,
    at={(ax1.east)}, anchor=west, xshift=2mm,
    title={Language-specific\\$\mathrm{LSS} \ge 0.5$, $n = 655$},
    yticklabels=\empty,
    colorbar horizontal,
    colorbar style={
        at={(-0.0333,-0.413)}, anchor=north,
        width=3.0cm, height=0.13cm,
        scaled x ticks=false,
        xtick={-0.003,0,0.003},
        xticklabels={$-0.003$,$0$,$0.003$},
        xlabel={Mean bias $\Delta$},
        xlabel style={font=\scriptsize, yshift=21pt, xshift=-70pt},
        tick label style={font=\scriptsize},
        tick style={draw=none},
        axis line style={draw=black!45, line width=0.4pt},
    },
]
    \addplot [matrix plot, mesh/cols=3, point meta=explicit] coordinates {
        (0,0) [0.001] (1,0) [-0.002] (2,0) [-0.001]
        (0,1) [0.001] (1,1) [-0.002] (2,1) [-0.001]
        (0,2) [0.001] (1,2) [-0.002] (2,2) [-0.001]
        (0,3) [0.001] (1,3) [-0.002] (2,3) [-0.001]
    };
\end{axis}
\end{tikzpicture}

%% file: assets/steering/llama_multi/02b_language_transfer_by_lss.tex
\begin{tikzpicture}

\pgfplotsset{
    colormap={diverging}{
        color(0cm)=(pblue!35!white);
        color(1cm)=(white);
        color(2cm)=(pred!35!white)
    },
}

\pgfplotsset{
    transfermatrix/.style={
        enlargelimits=false,
        axis on top,
        scale only axis,
        width=3.0cm,
        height=2cm,
        axis line style={draw=black!45, line width=0.4pt},
        tick style={draw=none},
        title style={align=center, font=\scriptsize, yshift=-4pt},
        label style={font=\scriptsize},
        tick label style={font=\scriptsize},
        xtick={0,1,2},
        xticklabels={\strut en, \strut es, \strut nl},
        ytick={0,1,2,3},
        point meta min=-0.001, point meta max=0.001,
        nodes near coords,
        nodes near coords style={
            anchor=center,
            font=\scriptsize,
            /pgf/number format/fixed,
            /pgf/number format/fixed zerofill,
            /pgf/number format/precision=3,
        },
    },
}

\begin{axis}[
    transfermatrix,
    name=ax1,
    title={Language-agnostic\\$\mathrm{LSS} < 0.5$, $n = 36$},
    yticklabels={\strut tr, \strut nl, \strut es, \strut en},
    ylabel={Feature language},
]
    \addplot [matrix plot, mesh/cols=3, point meta=explicit] coordinates {
        (0,0) [-0.000] (1,0) [0.000] (2,0) [0.000]
        (0,1) [-0.000] (1,1) [0.000] (2,1) [0.001]
        (0,2) [-0.001] (1,2) [0.000] (2,2) [0.000]
        (0,3) [-0.000] (1,3) [-0.000] (2,3) [0.001]
    };
\end{axis}

\begin{axis}[
    transfermatrix,
    name=ax2,
    at={(ax1.east)}, anchor=west, xshift=2mm,
    title={Language-specific\\$\mathrm{LSS} \ge 0.5$, $n = 459$},
    yticklabels=\empty,
    colorbar horizontal,
    colorbar style={
        at={(-0.0333,-0.413)}, anchor=north,
        width=3.0cm, height=0.13cm,
        scaled x ticks=false,
        xtick={-0.001,0,0.001},
        xticklabels={$-0.001$,$0$,$0.001$},
        xlabel={Mean bias $\Delta$},
        xlabel style={font=\scriptsize, yshift=21pt, xshift=-70pt},
        tick label style={font=\scriptsize},
        tick style={draw=none},
        axis line style={draw=black!45, line width=0.4pt},
    },
]
    \addplot [matrix plot, mesh/cols=3, point meta=explicit] coordinates {
        (0,0) [0.001] (1,0) [-0.001] (2,0) [0.000]
        (0,1) [0.001] (1,1) [-0.001] (2,1) [0.000]
        (0,2) [0.001] (1,2) [-0.001] (2,2) [0.000]
        (0,3) [0.001] (1,3) [-0.000] (2,3) [0.000]
    };
\end{axis}
\end{tikzpicture}

%% file: assets/steering/llama_scope/02d_category_transfer_by_lss_resid.tex
\begin{tikzpicture}

\pgfplotsset{
    colormap={diverging}{
        color(0cm)=(pblue!35!white);
        color(1cm)=(white);
        color(2cm)=(pred!35!white)
    },
}

\pgfplotsset{
    transfermatrix/.style={
        enlargelimits=false,
        axis on top,
        scale only axis,
        width=3.0cm,
        height=2.5cm,
        axis line style={draw=black!45, line width=0.4pt},
        tick style={draw=none},
        title style={align=center, font=\scriptsize, yshift=-4pt},
        label style={font=\scriptsize},
        tick label style={font=\scriptsize},
        xtick={0,1,2},
        xticklabels={\strut en, \strut es, \strut nl},
        ytick={0,1,2,3,4,5},
        point meta min=-0.12, point meta max=0.12,
        nodes near coords,
        nodes near coords style={
            anchor=center,
            font=\fontsize{5.5}{5.5}\selectfont,
            /pgf/number format/fixed,
            /pgf/number format/fixed zerofill,
            /pgf/number format/precision=4,
        },
    },
}

\begin{axis}[
    transfermatrix,
    name=ax1,
    title={Language-agnostic\\$\mathrm{LSS} < 0.5$, $n = 98$},
    yticklabels={\strut Sex.Or., \strut SES, \strut Appear., \strut Gender, \strut Disab., \strut Age},
    ylabel={Discovery category},
]
    \addplot [matrix plot, mesh/cols=3, point meta=explicit] coordinates {
        (0,0) [-0.1149] (1,0) [-0.0798] (2,0) [-0.0964]
        (0,1) [-0.1066] (1,1) [-0.0842] (2,1) [-0.0908]
        (0,2) [-0.1169] (1,2) [-0.0823] (2,2) [-0.0954]
        (0,3) [-0.1037] (1,3) [-0.0786] (2,3) [-0.0906]
        (0,4) [-0.1101] (1,4) [-0.0817] (2,4) [-0.0933]
        (0,5) [-0.1088] (1,5) [-0.0783] (2,5) [-0.0937]
    };
\end{axis}

\begin{axis}[
    transfermatrix,
    name=ax2,
    at={(ax1.east)}, anchor=west, xshift=2mm,
    title={Language-specific\\$\mathrm{LSS} \ge 0.5$, $n = 437$},
    yticklabels=\empty,
    colorbar horizontal,
    colorbar style={
        at={(-0.0333,-0.31)}, anchor=north,
        width=3.0cm, height=0.13cm,
        scaled x ticks=false,
        xtick={-0.12,0,0.12},
        xticklabels={$-0.12$,$0$,$0.12$},
        xlabel={Mean bias $\Delta$},
        xlabel style={font=\scriptsize, yshift=21pt, xshift=-70pt},
        tick label style={font=\scriptsize},
        tick style={draw=none},
        axis line style={draw=black!45, line width=0.4pt},
    },
]
    \addplot [matrix plot, mesh/cols=3, point meta=explicit] coordinates {
        (0,0) [-0.0356] (1,0) [-0.0026] (2,0) [-0.0781]
        (0,1) [-0.0070] (1,1) [0.0052] (2,1) [-0.0801]
        (0,2) [-0.0303] (1,2) [0.0015] (2,2) [-0.0770]
        (0,3) [-0.0176] (1,3) [-0.0004] (2,3) [-0.0824]
        (0,4) [-0.0334] (1,4) [-0.0035] (2,4) [-0.0829]
        (0,5) [-0.0266] (1,5) [-0.0009] (2,5) [-0.0791]
    };
\end{axis}
\end{tikzpicture}

%% file: assets/steering/gemma_scope/02d_category_transfer_by_lss_resid.tex
\begin{tikzpicture}

\pgfplotsset{
    colormap={diverging}{
        color(0cm)=(pblue!35!white);
        color(1cm)=(white);
        color(2cm)=(pred!35!white)
    },
}

\pgfplotsset{
    transfermatrix/.style={
        enlargelimits=false,
        axis on top,
        scale only axis,
        width=3.0cm,
        height=2.5cm,
        axis line style={draw=black!45, line width=0.4pt},
        tick style={draw=none},
        title style={align=center, font=\scriptsize, yshift=-4pt},
        label style={font=\scriptsize},
        tick label style={font=\scriptsize},
        xtick={0,1,2},
        xticklabels={\strut en, \strut es, \strut nl},
        ytick={0,1,2,3,4,5},
        point meta min=-0.07, point meta max=0.07,
        nodes near coords,
        nodes near coords style={
            anchor=center,
            font=\fontsize{5.5}{5.5}\selectfont,
            /pgf/number format/fixed,
            /pgf/number format/fixed zerofill,
            /pgf/number format/precision=4,
        },
    },
}

\begin{axis}[
    transfermatrix,
    name=ax1,
    title={Language-agnostic\\$\mathrm{LSS} < 0.5$, $n = 42$},
    yticklabels={\strut Sex.Or., \strut SES, \strut Appear., \strut Gender, \strut Disab., \strut Age},
    ylabel={Discovery category},
]
    \addplot [matrix plot, mesh/cols=3, point meta=explicit] coordinates {
        (0,0) [-0.0047] (1,0) [-0.0226] (2,0) [-0.0104]
        (0,1) [0.0047] (1,1) [0.0006] (2,1) [0.0101]
        (0,2) [0.0053] (1,2) [0.0007] (2,2) [0.0116]
        (0,3) [0.0065] (1,3) [-0.0004] (2,3) [0.0060]
        (0,4) [-0.0082] (1,4) [-0.0335] (2,4) [-0.0211]
        (0,5) [-0.0059] (1,5) [-0.0297] (2,5) [-0.0172]
    };
\end{axis}

\begin{axis}[
    transfermatrix,
    name=ax2,
    at={(ax1.east)}, anchor=west, xshift=2mm,
    title={Language-specific\\$\mathrm{LSS} \ge 0.5$, $n = 702$},
    yticklabels=\empty,
    colorbar horizontal,
    colorbar style={
        at={(-0.0333,-0.31)}, anchor=north,
        width=3.0cm, height=0.13cm,
        scaled x ticks=false,
        xtick={-0.07,0,0.07},
        xticklabels={$-0.07$,$0$,$0.07$},
        xlabel={Mean bias $\Delta$},
        xlabel style={font=\scriptsize, yshift=21pt, xshift=-70pt},
        tick label style={font=\scriptsize},
        tick style={draw=none},
        axis line style={draw=black!45, line width=0.4pt},
    },
]
    \addplot [matrix plot, mesh/cols=3, point meta=explicit] coordinates {
        (0,0) [0.0044] (1,0) [-0.0640] (2,0) [0.0236]
        (0,1) [0.0055] (1,1) [-0.0619] (2,1) [0.0264]
        (0,2) [0.0037] (1,2) [-0.0632] (2,2) [0.0241]
        (0,3) [0.0011] (1,3) [-0.0673] (2,3) [0.0156]
        (0,4) [0.0085] (1,4) [-0.0695] (2,4) [0.0193]
        (0,5) [0.0057] (1,5) [-0.0632] (2,5) [0.0223]
    };
\end{axis}
\end{tikzpicture}

%% file: assets/steering/qwen_multi/02d_category_transfer_by_lss.tex
\begin{tikzpicture}

\pgfplotsset{
    colormap={diverging}{
        color(0cm)=(pblue!35!white);
        color(1cm)=(white);
        color(2cm)=(pred!35!white)
    },
}

\pgfplotsset{
    transfermatrix/.style={
        enlargelimits=false,
        axis on top,
        scale only axis,
        width=3.0cm,
        height=2.5cm,
        axis line style={draw=black!45, line width=0.4pt},
        tick style={draw=none},
        title style={align=center, font=\scriptsize, yshift=-4pt},
        label style={font=\scriptsize},
        tick label style={font=\scriptsize},
        xtick={0,1,2},
        xticklabels={\strut en, \strut es, \strut nl},
        ytick={0,1,2,3,4,5},
        point meta min=-0.0055, point meta max=0.0055,
        nodes near coords,
        nodes near coords style={
            anchor=center,
            font=\fontsize{5.5}{5.5}\selectfont,
            /pgf/number format/fixed,
            /pgf/number format/fixed zerofill,
            /pgf/number format/precision=4,
        },
    },
}

\begin{axis}[
    transfermatrix,
    name=ax1,
    title={Language-agnostic\\$\mathrm{LSS} < 0.5$, $n = 44$},
    yticklabels={\strut Sex.Or., \strut SES, \strut Appear., \strut Gender, \strut Disab., \strut Age},
    ylabel={Discovery category},
]
    \addplot [matrix plot, mesh/cols=3, point meta=explicit] coordinates {
        (0,0) [0.0024] (1,0) [0.0021] (2,0) [0.0006]
        (0,1) [0.0043] (1,1) [0.0036] (2,1) [0.0037]
        (0,2) [0.0053] (1,2) [0.0054] (2,2) [0.0052]
        (0,3) [0.0032] (1,3) [0.0032] (2,3) [0.0034]
        (0,4) [0.0029] (1,4) [0.0036] (2,4) [0.0025]
        (0,5) [0.0019] (1,5) [0.0018] (2,5) [0.0012]
    };
\end{axis}

\begin{axis}[
    transfermatrix,
    name=ax2,
    at={(ax1.east)}, anchor=west, xshift=2mm,
    title={Language-specific\\$\mathrm{LSS} \ge 0.5$, $n = 655$},
    yticklabels=\empty,
    colorbar horizontal,
    colorbar style={
        at={(-0.0333,-0.31)}, anchor=north,
        width=3.0cm, height=0.13cm,
        scaled x ticks=false,
        xtick={-0.0055,0,0.0055},
        xticklabels={$-0.0055$,$0$,$0.0055$},
        xlabel={Mean bias $\Delta$},
        xlabel style={font=\scriptsize, yshift=21pt, xshift=-70pt},
        tick label style={font=\scriptsize},
        tick style={draw=none},
        axis line style={draw=black!45, line width=0.4pt},
    },
]
    \addplot [matrix plot, mesh/cols=3, point meta=explicit] coordinates {
        (0,0) [-0.0015] (1,0) [0.0008] (2,0) [-0.0014]
        (0,1) [-0.0011] (1,1) [0.0007] (2,1) [-0.0019]
        (0,2) [-0.0016] (1,2) [0.0007] (2,2) [-0.0013]
        (0,3) [-0.0018] (1,3) [0.0009] (2,3) [-0.0019]
        (0,4) [-0.0012] (1,4) [0.0007] (2,4) [-0.0018]
        (0,5) [-0.0014] (1,5) [0.0011] (2,5) [-0.0016]
    };
\end{axis}
\end{tikzpicture}

%% file: assets/steering/llama_multi/02d_category_transfer_by_lss.tex
\begin{tikzpicture}

\pgfplotsset{
    colormap={diverging}{
        color(0cm)=(pblue!35!white);
        color(1cm)=(white);
        color(2cm)=(pred!35!white)
    },
}

\pgfplotsset{
    transfermatrix/.style={
        enlargelimits=false,
        axis on top,
        scale only axis,
        width=3.0cm,
        height=2.5cm,
        axis line style={draw=black!45, line width=0.4pt},
        tick style={draw=none},
        title style={align=center, font=\scriptsize, yshift=-4pt},
        label style={font=\scriptsize},
        tick label style={font=\scriptsize},
        xtick={0,1,2},
        xticklabels={\strut en, \strut es, \strut nl},
        ytick={0,1,2,3,4,5},
        point meta min=-0.002, point meta max=0.002,
        nodes near coords,
        nodes near coords style={
            anchor=center,
            font=\fontsize{5.5}{5.5}\selectfont,
            /pgf/number format/fixed,
            /pgf/number format/fixed zerofill,
            /pgf/number format/precision=4,
        },
    },
}

\begin{axis}[
    transfermatrix,
    name=ax1,
    title={Language-agnostic\\$\mathrm{LSS} < 0.5$, $n = 36$},
    yticklabels={\strut Sex.Or., \strut SES, \strut Appear., \strut Gender, \strut Disab., \strut Age},
    ylabel={Discovery category},
]
    \addplot [matrix plot, mesh/cols=3, point meta=explicit] coordinates {
        (0,0) [0.0007] (1,0) [0.0001] (2,0) [0.0002]
        (0,1) [0.0012] (1,1) [0.0002] (2,1) [0.0002]
        (0,2) [-0.0003] (1,2) [-0.0001] (2,2) [-0.0001]
        (0,3) [0.0003] (1,3) [0.0002] (2,3) [-0.0004]
        (0,4) [0.0018] (1,4) [0.0008] (2,4) [0.0016]
        (0,5) [0.0006] (1,5) [-0.0001] (2,5) [0.0007]
    };
\end{axis}

\begin{axis}[
    transfermatrix,
    name=ax2,
    at={(ax1.east)}, anchor=west, xshift=2mm,
    title={Language-specific\\$\mathrm{LSS} \ge 0.5$, $n = 459$},
    yticklabels=\empty,
    colorbar horizontal,
    colorbar style={
        at={(-0.0333,-0.31)}, anchor=north,
        width=3.0cm, height=0.13cm,
        scaled x ticks=false,
        xtick={-0.002,0,0.002},
        xticklabels={$-0.002$,$0$,$0.002$},
        xlabel={Mean bias $\Delta$},
        xlabel style={font=\scriptsize, yshift=21pt, xshift=-70pt},
        tick label style={font=\scriptsize},
        tick style={draw=none},
        axis line style={draw=black!45, line width=0.4pt},
    },
]
    \addplot [matrix plot, mesh/cols=3, point meta=explicit] coordinates {
        (0,0) [0.0001] (1,0) [-0.0005] (2,0) [-0.0004]
        (0,1) [0.0006] (1,1) [-0.0007] (2,1) [-0.0002]
        (0,2) [0.0007] (1,2) [-0.0007] (2,2) [0.0002]
        (0,3) [0.0006] (1,3) [-0.0009] (2,3) [-0.0001]
        (0,4) [0.0007] (1,4) [-0.0003] (2,4) [-0.0002]
        (0,5) [0.0009] (1,5) [-0.0001] (2,5) [-0.0001]
    };
\end{axis}
\end{tikzpicture}